\documentclass[preprint,12pt]{elsarticle}

\usepackage{amssymb}
\usepackage{amsmath}

\journal{}

\begin{document}

\begin{frontmatter}



\title{Data-driven prediction of Air Traffic Controllers reactions to resolving conflicts.}


\author[inst1]{Alevizos Bastas}

\affiliation[inst1]{organization={Department of Digital Systems, University of Piraeus},
            country={Greece}}

\author[inst1]{George Vouros}


\begin{abstract}
With the aim to enhance automation in conflict detection and resolution (CD\&R) tasks in the Air Traffic Management domain, in this paper we propose deep learning techniques (DL) that can learn models of Air Traffic Controllers' (ATCO) reactions in resolving conflicts that can violate separation minimum constraints among aircraft trajectories: This implies learning $when$ the ATCO will react towards resolving a conflict, and $how$ he/she will react. Timely reactions, to which this paper aims, focus on $when$ do reactions happen, aiming to predict the trajectory points, as the trajectory evolves, that the ATCO issues a conflict resolution action, while also predicting the type of resolution action (if any). Towards this goal, the paper formulates the ATCO reactions prediction problem for CD\&R, and presents DL methods that can model ATCO timely reactions and evaluates these methods in real-world data sets, showing their efficacy in prediction with very high accuracy.
\end{abstract}



\begin{keyword}
Imitation Learning \sep Deep Learning \sep Variational AutoEncoders \sep Air Traffic Control \sep Conflict Detection and Resolution
\end{keyword}

\end{frontmatter}


\section{Introduction}
The Air Traffic Management (ATM) system has reached its limits regarding efficiency and cost effectiveness. This was evident in the pre COVID era and is expected to evolve further  in the post COVID era. To overcome the implied limitations, different initiatives world-wide such as NextGen in the US and SESAR in Europe have been exploring the use of automation and Artificial Intelligence to enable a more efficient ATM system.

This work aims to contribute to conflict detection and resolution (CD\&R) tasks executed as part of the Air Traffic Control (ATC\footnote{https://ext.eurocontrol.int/lexicon/index.php/Air\_traffic\_control}) service,  promoting safe, orderly and expeditious flow of air traffic, at the tactical phase of operations by modeling Air Traffic Controllers' (ATCO) reactions in resolving conflicts:
We propose data-driven, deep learning techniques to model ATCO reactions. In general, this implies  learning $when$ the ATCO will react to resolve a detected conflict, and $how$ he/she will react. Timely reactions, to which this paper aims, focus on $when$ do reactions happen, aiming to predict the trajectory points, as the trajectory evolves, that the ATCO issues a conflict resolution action, while also predicting the type of the conflict resolution action.
In so doing, we formulate the {\em ATCO reaction prediction problem} and present a data-driven deep learning method that models timely reactions.  This paves the way to (a) automate the ATC process, (b) optimizing the ATC process,  and ensuring enhanced decision making for ATCO, while leveraging
(c) human-AI collaboration in the context of ATC. 

Casting CD\&R as a data-driven problem from the perspective of modeling ATCO reactions, is novel  since, as far as we know, there is not any other study that does so. Indeed, most of the CD\&R approaches are trained, validated and tested in simulated settings. We conjecture that in safety critical domains such as ATC,  the actions proposed by automated systems should be  ``close'' to those taken by humans: By ``close'' we mean with a short distance in any of the temporal, spatial dimensions at which actions are decided and applied, as well as  w.r.t. action specific dimensions, such as changes in speed magnitude, temporal extent of manoeuvres, etc. This implies safety in the automation process, taking into account human expertise, (human-like) flexibility and tolerance in reacting to situations. We believe that such an approach,  is capable of developing trust to an automated system: Actions that are close to the human rationale are more understandable or self explanatory to human operators, and the system objectives can be made  intuitively transparent, given that the system models the ATCO objectives and preferences. 
Additionally, the automation of the CD\&R process at the tactical phase is expected to reduce the ATCO workload, increase the use of airspace, and minimize costs of operations. 


For such a data-driven imitation process historical expert samples (i.e. flown trajectories annotated with ATCO resolution actions) must  ideally indicate, together with the resolution actions, the observations perceived by ATCO before the resolution action, and which drove the specific action. Such observations regard the prediction of the evolution of the trajectories, and the assessment of conflicts. However, historical data sets indicate only the effect of ATCO resolution actions, but not the rationale behind them. This adds difficulty in the learning process, since imitating the $when$ and $how$ of the ATCO reactions necessitates recovering the state, or the important observations, that the ATCO faced or assessed, driving the decision. This is not a trivial task as the evolution of the trajectories is uncertain, and this uncertainty should be incorporated into the CD\&R process. Towards this goal, in this paper, we propose a  methodology for simulating the uncertainty in the trajectory evolution process towards revealing the ATCO observations that have triggered the reactions.
    
    The specific contributions made in this paper are as follows:
    
        -- We formulate the problem of learning the ATCO reactions;
        
        -- We propose using a supervised deep learning method employing  a Variational Auto-Encoder (VAE) for predicting ATCO reactions in a hierarchical manner, in the context of a methodology to model ATCO behaviour;
        
        -- We propose a data-driven method for simulating the uncertainty in the evolution of trajectories, revealing the rationale behind ATCO reactions;
        
        -- We propose a methodology for evaluating data-driven methods to resolve the ATCO reaction problem;
        
        -- We  evaluate the proposed method using real world data.

    The paper is structured as follows: Section \ref{sec:background} presents background knowledge, section \ref{sec:ProblemSpec} specifies the problem of predicting the ATCO reactions and section \ref{sec:solution} presents the proposed methodology. Section \ref{sec:experimental evaluation} presents the proposed evaluation methodology and experimental results, section \ref{sec:related work} presents related work, and section \ref{sec:concluding remarks} concludes the paper and presents future work.

\section{Background Knowledge}
\label{sec:background}

\subsection{Conflict Detection and Resolution}
\label{sec:CDR}

To maintain the risk of collision between aircraft in acceptable levels, the ATM system requires that the aircraft do not breach certain separation minima both at the horizontal and vertical axes. According to ICAO Doc 4444 the minimum prescribed horizontal separation when using surveillance systems is 5NM. This may be further reduced or increased by the Air Traffic Service (ATS) authority based on the surveillance systems' capabilities and the situation created between the aircraft. According to ICAO documents, the specified minimum vertical separation for Intrument Flight Rules (IFR) flights is 1000 ft (300 m) below FL290 and 2000 ft (600 m) from FL290 and above. When Reduced Vertical Separation Minima (RVSM) apply, this changes to 1000 ft (300 m) below FL410 and 2000 ft (600 m) from FL410 and above.\footnote{https://www.skybrary.aero/index.php/Separation\_Standards} A \textit{loss of separation} is defined as the violation of separation minima in controlled airspaces, whereas a \textit{conflict} is defined as a {\em predicted} violation of the separation minima.
 
 Nowadays conflicts are detected and resolved by the Planner Controller (PC) and the Executive Controller (EC). Tactical conflict detection and resolution is executed by  ECs detecting and resolving conflicts in their respective Areas of Responsibilities (AoRs)  (airspace sectors that segregate the airspace),  also coordinating actions with the ECs of the downstream sectors.  In contrast to that, which happens today, in a flight-centric ATC we may ignore AoRs  corresponding to sectors: Conflicts are detected in a temporal and spatial granularity which is larger than that of flights in sectors.

 
 Conflict detection and resolution in the planning phase  suggests changes in the flight plan.  At the tactical phase it implies changes of the actual flight trajectory, given the trajectory up to the current time point, the current flight plan, and/or prediction(s) on the evolution of the trajectory from that time point and on. Prediction is crucial here, since the future position of the aircraft is uncertain and the uncertainty grows larger with longer time horizons, limiting the confidence on predictions. This, combined to the (uncertain) evolution of other trajectories, implies uncertainties in conflicts that ATCO have to assess towards prescribing resolution actions.

  CD\&R involves human expertise and informed judgement. Thus, it is very difficult to hand-craft criteria which will drive  a system to decide whether a conflict deserves a certain reaction at a particular time point,  which conflict should be resolved among several co-occurring ones, and at which time point during the evolution of the involved trajectories one should react to a conflict, especially in long-term horizons (i.e. beyond 15-20 minutes). In addition, this task requires advanced trajectory prediction abilities, considering also the uncertainty in the evolution of trajectories.

\subsection{Imitation Learning}
\label{sec:imitation}

Imitation learning aims at learning policies that mimic expert behaviour from
demonstrations. Modeling the problem as a Markov Decision Process (MDP), the goal in imitation
learning is to learn a policy $\pi(a|s)$, which defines the conditional distribution over actions $a \in A$ for
any state $s$ of interest, given state-action sequences $T_E =  \{T|T=((s_0, a_0), \dots, (s_{|T|},a_{|T|}) )$ demonstrated by experts. 

As any imitation problem, the data-driven CD\&R  task can be defined as follows: Given a set $T_E$ of historical, demonstrated trajectories with conflict resolution actions (i.e., demonstrating ATCO actions  in resolving conflicts in the course of trajectory evolution),  the objective is to determine a policy $\pi$ and a “reward” function $R_E$ that determines the generation of maximal-expected-cumulative-reward in any trajectory $T_\pi$: 

\begin{center}
    $T_\pi=argmax_\pi \mathbb{E}_\pi [R_E (s,a)]$
\end{center}

The policy prescribes the probability of applying an action $a$ at  state $s$, so as the trajectory to evolve from that state on, maximizing the expected reward. In the context of CD\&R problem, the policy must ``shape" the trajectory w.r.t. the demonstrated expert behaviour in $T_E$, so as to resolve conflicts. Actions executed at each state determine how the trajectory evolves towards the next state (e.g. by means of change in speed, change in aircraft course, other detailed aircraft intent instructions etc.). 

Following a data-driven approach, the reward function shows the adherence of predictions to behavioural patterns and policies,  as these are demonstrated in historical cases $T_E$, and to additional constraints.
Many imitation learning methods do not require learning or crafting manually a reward function and this is the approach we follow in this work.

Bellow we provide succinct information on state of the art imitation learning methods for trajectories, in order to motivate and explain better the proposal made in this paper and the methodology followed.

Ho et. al, in 
\cite{ho2016generative} introduced an imitation learning framework called Generative Adversarial
Imitation Learning (GAIL) that is able to learn policies for complex high-dimensional physics-based
control tasks. 



While GAIL can learn a policy that imitates the evolution of demonstrated trajectories,  considering CD\&R tasks as sub tasks that must be executed while a flight trajectory evolves, we aim at imitating intra-trajectory (i.e., intra-flight) modes of behaviour, that GAIL cannot detect.  Directed InfoGAIL proposed in \cite{sharma2018directed} aims to model such intra-trajectory sub-tasks' variations by means of latent codes. Latent codes correspond to sub-tasks variations within a demonstration and can be considered in a hierarchical manner as high-level actions ``shaping a trajectory", refined by primitive or low level policy $\pi$ actions. 




In a more technical level, Directed InfoGAIL \cite{sharma2018directed}, in contrast to other approaches, forces the policy to generate trajectories that maximize the directed or
causal information flow from trajectories to the sequence of latent variables: 
Given a trajectory T generated up to current time $t$, $T^{1:t}=(s_1, \dots , a_{t-1}, s_t)$, Directed InfoGAIL derives the variational lower bound 
$L_1(\pi, q)$ of the mutual information $I(c; T )$ between latent codes and trajectories, by using 
an approximated posterior $q(c^t|c^{1:t-1}, T^{1:t-1})$ for the mode to follow the sequence of modes up to current time point $t-1$, and $T^{1:t-1}$, instead of the true posterior $p(c^t|c^{1:t-1}, T^{1:|T|})$:
\small
\[L_1(\pi, q) = \sum_t \mathbb{E}_{ c^{1:t} \sim p(c1:t), a^{t-1}  \sim  \pi(\cdot |s^{t-1},c^{1:t-1})}log  (q(c^t|c^{1:t-1},T^{1:t})) +H(c) \leq I(c;T)\]
\normalsize
Thus,  at each time point $t$,
this method learns a posterior
distribution over the latent code $c$ at $t$, given the latent factors discovered up to $t-1$, and the trajectory up
to $t$.
To estimate this distribution, Directed InfoGAIL trains a
variational auto-encoder (VAE) \cite{kingma2013auto} on the expert trajectories.

Following the above methodology to model the ATCO policy in resolving conflicts, latent codes represent high-level ATCO reactions to detected conflicts, which are further refined by low-level conflict resolution  actions that evolve the flight trajectory. 
In this paper we report on what we consider the first stage of this methodology: i.e. the prediction of ATCO reactions at any specific trajectory point. We do this by training a model in a supervised way, exploiting demonstrated ATCO resolution actions associated to historical  conflict-free trajectories,  estimating the features that affect ATCO decisions.


\subsection{Variational Auto-encoders (VAE)}
\label{sec:VAE}

This section provides background knowledge and preliminaries regarding VAE, which is the model that we use in the first methodological stage.

Auto-encoders are neural network models trained to reconstruct the input to their output. Internally they can be broken down into two parts: an encoder network and a decoder network. The encoder network, comprising a number of hidden layers, maps the input $x$ to an encoding $c$, which can be denoted as $c = q_\phi(x)$, where $\phi$ are the parameters of the encoder network. The decoder network maps the encoding $c$ to a reconstruction of the input $r = p_\theta(c)$, where $\theta$ are the decoder parameters. Auto-encoders do not merely learn to reconstruct the input. They learn an encoded representation $c$ of the input $x$, which retains enough information to allow a reconstruction $r$ (e.g. for dimensionality reduction or feature learning in \cite{AEkramer1991nonlinear}, \cite{hinton2006reducing}).
Auto-encoders have been explored for more than three decades, with recent advances applying auto-encoders to image de-noising \cite{xie2012image}, anomaly detection \cite{sakurada2014anomaly}, information retrieval \cite{krizhevsky2011using} and generative tasks (i.e. image captioning \cite{pu2016variational}).

Variational Auto-encoders (VAE) \cite{kingma2013auto} are a generative variant of auto-encoder models, successfully applied to generative tasks: 
The encoder of a variational auto-encoder outputs the parameters of a distribution $q_\phi(c|x)$ approximating the true intractable posterior $p(c|x)$. The decoder samples $c$ from $q_{\phi}(c|x)$ and outputs a reconstruction $r$.
Therefore, VAE learns an approximation $q_\phi(c|x)$ of the true intractable posterior $p(c|x)$, represented by the encoder, and a generative model $p_\theta(r|c)$, represented by the decoder. To do so VAE maximizes the lower bound:
\begin{equation}
    L(\theta,\phi;x)=-D_{KL}(q_\phi(c|x)||p(c))+\mathbb{E}_{q_\phi(c|x)}log(p_\theta(x|c))
\end{equation}

In the context of Directed InfoGAIL \cite{sharma2018directed} the encoder approximates the true posterior $p(c^t|c^{1:t-1}, T^{1:|T|})$, predicting the latent codes while performing tasks. The decoder learns a policy, generating actions, given the state and the predicted latent code, according to the demonstrated examples. To do so, the VAE used minimizes the following objective:
\begin{center}
    $L_{VAE} = -\sum\limits_{t}\mathop{\mathbb{E}}_{c^{t} \sim q}[log\pi(a^t|s^t,c^{1:t})] + \sum\limits_{t}D_{KL}(q(c^t|c^{1:t-1},\tau^{1:t})||p(c^t|c^{1:t-1}))$
\end{center}

Therefore, the encoder  is trained w.r.t. the errors propagating backwards from the decoder, following a hierarchical approach. Indeed, the  variational auto-encoder provides a hierarchical structure, where the encoder predicts the mode of behaviour $c$ (high-level actions), and the decoder predicts the policy (low-level) actions $\pi(a|s,c)$, given the state $s$ and the predicted $c$. 
Motivated by the Hierarchical Reinforcement Learning literature \cite{dietterich2000hierarchical}, \cite{sutton1999between}, \cite{kulkarni2016hierarchical} we consider a hierarchical structure of ATCO reactions where abstract high-level reactions, corresponding to modes of the ATCO behavior (indicating whether to issue a resolution action in the presence of a conflict) are further refined by means of fine low-level reactions that imply the evolution of aircraft state in a specific manner. 


\section{Problem Specification}
\label{sec:ProblemSpec}

\subsection{Definitions}
A trajectory is a chronologically ordered sequence
of states, without an explicit  consideration on actions shaping the trajectory:
$T=(s_0,s_1,... s_{|T|})$.

In the aviation domain, most relevant state variables are airspeeds, 3D position (determined by latitude (f), longitude (l) and geodetic altitude (h)), and time (t). The direction of a trajectory at any point is usually expressed in degrees from North (true, magnetic or grid).
Here, we aim to exploit 
aircraft trajectories whose states include 3D aircraft position with timestamps, in conjunction to contextual features that are useful to the CD\&R task.
Adding contextual features in a trajectory state results in a trajectory with {\em enriched points} or {\em enriched states}, thus to an {\em enriched trajectory}:

An {\em enriched  trajectory state} or {\em enriched  trajectory point} of a trajectory of length $|T|$, is defined to be a triplet $s_{r,i}=\langle p_i, t_i, v_i\rangle$, where $p_i$ is a point in the 3D space,  $v_i$ is a vector consisting of categorical and/or numerical variables, and $t_i$ is a timestamp, with $i\in [0,|T|-1$]. An {\em enriched trajectory} $T$ is defined to be a sequence of enriched states $s_{r,i}=\langle p_i, t_i, v_i\rangle$, $i\in [0,|T|-1]$.

A predicted trajectory $T_p$, is defined to be a specification of the future evolution of the aircraft state as a function of (a) the current flight conditions (e.g. an initial state with co-occurring trajectories according to flight plans or other predictions, actual weather conditions etc.), (b) a forecast of contextual features (e.g. forecast of weather conditions at specific points/regions, or predicted states of other flights) and (c) a ``policy" on how the trajectory evolves, i.e. a specification of how the aircraft is to transit among subsequent states starting from an initial state and on.

Given the evolution of a trajectory $T^{1:t}$, or simply $T^{t}$,  up to $t$, a predicted trajectory from that time point will be denoted as $T^t_p$. A set of such predictions, showing potential trajectory evolutions of $T^t$, is denoted by $\textbf{T}^t_p$.

The Closest Point of Approach (CPA)\footnote{https://www.skybrary.aero/index.php/Closest\_Point\_of\_Approach\_(CPA)} of an aircraft $i$ w.r.t. another aircraft $j$, is the position of $i$ when the distance between the two aircraft is minimum. The CPA can be computed in the horizontal axis, in the vertical axis or in all 3 dimensions. The time at CPA is the time at which the smallest distance between the two aircraft occurs.
To compute the CPA at the horizontal axis we follow the methodology presented in \cite{pham2019machine}. 

The Crossing Point (CP) \footnote{https://www.skybrary.aero/index.php/Vectoring\_Geometry} of a pair of aircraft $\langle i, j \rangle$ is the point at which the tracks of the aircraft intersect. Where the track\footnote{https://www.skybrary.aero/index.php/Heading,\_Track\_and\_Radial} of an aircraft is defined as the projection of the aircraft trajectory on the earth’s surface. 

Given a spatiotemporal area $SA$, \textit{neighbouring trajectories in SA} are those trajectories that co-occur in $SA$, i.e. with temporally corresponding points in $SA$ (3D points with equal timestamps),  satisfying also constraints regarding their tracks, CPA and CP. 

More formally:
\\

$Neigh(SA,t)=\{\{T_i, T_j\}| \textit{There is at least one point } (s_i,t) \textit{ in } T_i  \textit{ and one}\\ \textit{point } (s_j,t) \textit{ in } T_j, \textit{s.t. it holds that } in(s_i, SA) \textit{ and } in(s_j,SA) \textit{ at time point } t \\\textit{and also the aircraft flying } T^t_i \textit{ and } T^t_j  \textit{satisfy a} \textit{ set of constraints CR.}$
\\


The predicate $in(s,SA)$ is true when the 3D spatial point corresponding to $s$ is in the spatial region $SA$.
The set of constraints $CR$ includes the following:

- Aircraft have not crossed the crossing point;

- The tracks of the aircraft cross in less than $ct_{th}$ minutes;

- The horizontal distance at the CPA is less than $cpa_{d_{h_{th}}}nm$,

- The time to the CPA is less than $cpa_{t_{th}}min$;

- Aircraft altitude difference at the current time point is less than $d_{v_{th}} ft$.

 \noindent The parameter $ct_{th}$ is the crossing time threshold, $cpa_{d_{h_{th}}}$ is the horizontal distance threshold at the CPA, $cpa_{t_{th}}$ is the time to CPA threshold, and $d_{v_{th}}$ is the vertical distance threshold. We set $ct_{th}=20$ minutes, $cpa_{d_{h_{th}}}=15$ nm, $cpa_{t_{th}}=30$ minutes and $d_{v_{th}}$ to the vertical separation minimum (1000 feet under FL410 and 2000 feet from FL410 and over).

We consider aircraft flying trajectories in $Neigh(SA,t)$, to be in conflict. 

It must be noted that we use a large horizontal distance threshold at the CPA ($cpa_{d_{h_{th}}}$) of 15nm, in order to include margins of error when estimating ATCO observations in triggering their reactions. In so doing we incorporate further uncertainties in detecting conflicts. Indeed, ATCO take margins of error perceiving flights,\footnote{These margins may not be so large as we assume here but it must be emphasized that we do so in order to reveal the situation that ATCO perceive prior to their reaction.} even when the predicted horizontal distance between them at the CPA is greater than the horizontal separation minimum. Such margins of error are important to ensure safety.

In addition, given a specific (focal/own) trajectory $T_f$ and a spatiotemporal area $SA$, we define the set of \textit{neighboring} and thus \textit{conflicting trajectories to $T_f$ in $SA$ at a specific time point t}, or the set of trajectories interacting with $T_f$ in $SA$ at time point $t$, denoted $Neigh(T_f,SA,t)$, those that
a) have at least one point spatially close to the focal trajectory point at the time instance $t$, according to a horizontal distance measure $horizontal\_distance$ and a distance threshold $D_{th}$, and
b) satisfy the constraints  $CR$.

Formally:
\\

$Neigh(T_f, SA, t)=\{T| \textit{There is a point } (s_i,t) \textit{ in } T  \textit{ and a point } (s_j,t) \textit{ in }\\ T_f, \textit{ s.t. it holds that } in(s_i, SA) \textit{ and } in(s_j,SA) \textit{ at time point } t , \\ horizontal\_distance(s_i,s_j) \leq D_{th} \textit{ and the aircraft flying } T^t_f \textit{ and } T^t \textit{ satisfy}\\ \textit{the set of constraints in CR}\}.$
\\

\subsection{ATCO Reaction Prediction Problem Specification}
\label{sec:formal specification}

Given a set $\textbf{T}_E$ of historical trajectories and a  set  $\textbf{RA}_E$ of historical  ATCO conflict resolution actions associated to  trajectories in $\textbf{T}_E$, our goal is to learn a model that imitates the ATCO's behaviour in terms of these resolution actions. 
Considering a set of ATCO behavior modes abstracting resolution actions assigned to trajectories when conflicts are detected, our goal is to predict \textit{whether}, \textit{when} and \textit{how} the ATCO will react in assigning a conflict resolution action. 


So, in order to be able to imitate the behaviour of the ATCO and also the evolution of the trajectories in time, given $\textbf{T}_E$ and $\textbf{RA}_E$, we develop models that:
\begin{enumerate}
    \item Predict at any trajectory point the ATCO mode of behaviour deciding whether the ATCO would issue a resolution action, and 
	\item Predict the resolution action the ATCO would decide, if any.
\end{enumerate}
In this paper, focusing on ATCO reactions, we mostly focus on the 1st problem. In this case, modes of the ATCO behavior, in the more abstract form, can include: ``Not Assigning resolution action'' and ``Assigning resolution action''. Thus, we consider modes of ATCO behaviour as decisions corresponding to \textit{when}, i.e., at which points of the trajectory, the ATCO  issues a resolution action.  However, we conjecture that developing models of ATCO timely reactions, we need to also consider the 2nd problem. Thus,  models 
are trained to predict high-level reactions corresponding to modes, as well as low-level ATCO resolution actions. Here we focus on the 1st problem. Learning the low-level ATCO policy is a problem that will be addressed more adequately in future works, according to the methodology followed.




Given the above, the \textit{ATCO Reaction Prediction Problem} 
is about  predicting at a time point $t$ $whether$, $when$ and $how$ the ATCO will react  regarding a particular flight  that has executed  a trajectory $T_f^{t}$ (focal trajectory) given $Neigh(T_f,SA,t)$, w.r.t. a spatial area of responsibility $SA$. 

This problem  comprises (a) detecting conflicts by identifying neighbouring trajectories in the spatio-temporal region $SA$, 
 (b) determining the exact time point $t_A$, s.t. $t \leq t_A \leq t_c$ for issuing a resolution action, if any; where $t_c$ is the time point at which the conflict occurs, and (c)
deciding the resolution action to be applied, thus shaping the future evolution of the trajectory. 


It must be noted that, given multiple aircraft executing neighbouring trajectories, we also need to consider the problem of deciding which of the involved aircraft must manoeuvre to resolve any such conflict. This problem will be addressed in future work, together with learning the ATCO low-level policy on resolution actions.

\section{Solving the  ATCO Reaction Problem}
\label{sec:solution}

\begin{figure}[h]
\centering
\includegraphics[width=0.7 \linewidth]{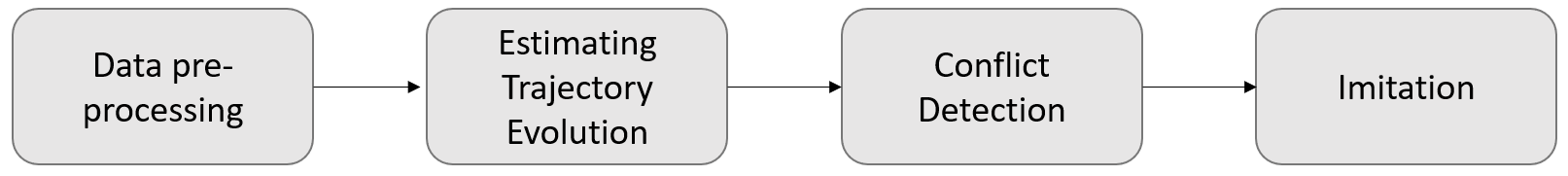}
\caption{Methodology stages for solving the ATCO reaction problem}
\label{fig:methodology}
\end{figure}


    
In this section we describe the methodology proposed for the ATCO reaction prediction problem. Figure \ref{fig:methodology} depicts the  stages of this methodology, and subsequent subsections describe each stage in detail, starting with the available data sources and their association. Briefly:

At the \textit{data pre-processing} stage we  pre-process and associate the available data sources of historical flown, thus conflict-free, trajectories, and ATCO events.

At the \textit{estimating trajectory evolution} stage we  estimate for any trajectory $T^t$, and at any time point $t$, the set of potential trajectory evolutions $\textbf{T}^t_p$, modelling the uncertainty in the trajectory evolution that ATCO somehow estimate towards detecting conflicts.

The conflict detection stage exploits the predicted evolutions $\textbf{T}^t_p$ of trajectories to identify neighbouring trajectories within an area of responsibility, and compute relevant features.

Finally, having detected conflicts that may have affected the evolution of the trajectories, and associated ATCO events, at the \textit{imitation stage} we proceed to training our models to imitate ATCO reactions.

\subsection{Data Sources}
Data sources comprise (a) surveillance data (IFS) of operational quality data with actual flights' trajectories (Spanish ATC Platform SACTA),
and (b) ATCO events that provide information regarding resolution actions assigned to flights by the ATCO (ATON, Automated NORVASE Takes).
	
Surveillance data include radar track points with temporal distance of approximately 5 seconds. At the pre-processing phase we synchronize all flights in the temporal dimension by interpolating  trajectory points at time points with a value multiple of 5 seconds. 


The surveillance dataset provides the spatiotemporal points of the trajectories, specifically the position (longitude, latitude, altitude) and the timestamp together with other information that identifies a specific trajectory, such as the callsign and the origin and destination airports. Such information is essential in order to associate a trajectory with ATCO events.  Given the surveillance data set, an area $SA$ and a particular flight we can determine at any time point $t$ the trajectory $T^t$ within $SA$, i.e. the trajectory up to that time point, as well as $Neigh(T^t,SA,t)$.

The ATCO events dataset provides information regarding conflict resolution actions issued by ATCO. It provides the callsign of the trajectory, the origin and destination airports, the timestamp of the resolution action and the type of the resolution action: $\langle \textit{callsign,\ origin\ airport,\ destination\ air-}\\ \textit{port,\ resolution\ action\ type}\rangle$\footnote{In this work we are interested only on the type of the resolution action, as the computation of the ATCO policy for the exact resolution  is out of scope.}. This information enables the association of ATCO conflict resolution actions with 
trajectory points.

Specifically, we associate an ATCO event for  a resolution action RA to a trajectory T when the following conditions are satisfied:
\begin{enumerate}
    \item RA.callsign = T.callsign
    \item RA.departure\_airport = T.departure\_airport
    \item RA.destination\_airport = T.destination\_airport
    \item Timestamp of the first T point $\leq$ timestamp of RA $\leq$ timestamp of the last T point.
\end{enumerate}

Given that the above conditions hold T and RA, the trajectory point that is temporally closer to the RA, is associated with the RA.

\begin{figure}[h]
\centering
\includegraphics[width=0.7\linewidth]{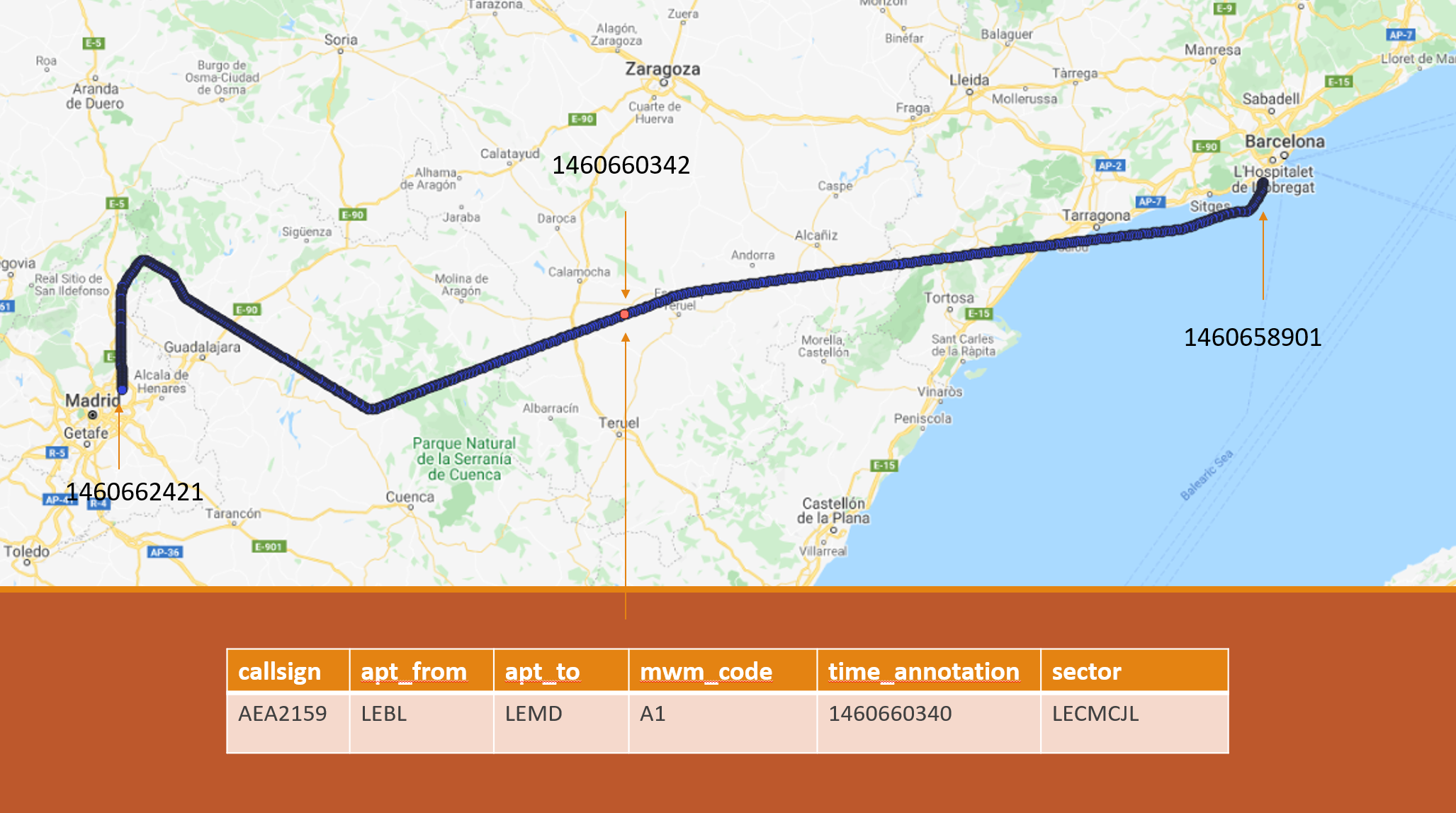}
\caption{Trajectory points (blue points) associated with a corresponding ATCO event. The figure indicates the callsign, the departure (apt\_from) and the destination airports (apt\_to), the resolution action type  (mwm\_code), the time (time\_annotation) and the sector in which the resolution was issued (sector). The red point depicts the  trajectory point  (with the timestamp) associated to the ATCO event.}
\label{fig:DataAssociationExample}
\end{figure}

Figure \ref{fig:DataAssociationExample} depicts an example of associating an ATCO event to the corresponding  trajectory. The table reports the attributes of the ATCO event, the blue line is the trajectory and the red point depicts the  trajectory point which is closer in time to the ATCO resolution action, together with its timestamp. The  timestamps of the ATCO event, of the first and of the last point of the trajectory, are provided.



\subsection{Simulating uncertainty in trajectory evolution}
\label{sec:simulating uncertainty}

Addressing the ATCO reaction prediction problem in a data-driven way (i.e. based on historical data), the training process necessitates having data associating ATCO observations that drove estimations for conflicts, with specific reactions. These observations are  not recorded in a historical data set and they  concern contextual features of trajectories, revealing  ATCO’s estimations on the future evolution of trajectories at any time point, and conflicts with neighboring trajectories, prior to implementing a resolution action. These observations must enrich the states of a trajectory and be exploited during the training process.

Determining these observations, it involves detecting the conflicts that would occur if the ATCO would not react at a time point $t$. It involves estimating the cases where two trajectories $T_1^t$ and $T_2^t$ are in conflict and violate the separation minima according to assessed trajectories' evolutions $T_{1p}^t$ and $T_{2p}^t$.
This is not trivial as the state of the aircraft after $t$ is uncertain, and this uncertainty grows exponentially as the temporal horizon of estimation increases.

To estimate the evolution  of trajectories we exploit $\Delta c$ and $\Delta s_h$ of the historical trajectories. $\Delta c$ is the difference $c_{t+1}-c_t$ where $c_t$ is the aircraft's course at time point $t$ and $\Delta s_h$ is the difference $s_{h_{t+1}} - s_{h_{t}}$ where $s_{h_{t}}$ is the magnitude of the horizontal speed at time point $t$.



More specifically, given trajectories $T_E$ in the surveillance data set, at each time point we compute $\Delta c$ and $\Delta s_h$ and we  divide the values in $n$ equi-height bins, where $n$ is a hyper-parameter which we set to 20 in our experiments. Using the median of each bin as the bin's representative value,  we get $n$ values for the potential deviation in the course and $n$ values for the potential deviation in the horizontal speed.

Overall, given a  trajectory T up to the current time point $t$, i.e., $T^t$, we compute the course ($course^t$) that the aircraft follows and the magnitude of the aircraft’s speed ($speed^t$) at that  point $s_t$, exploiting information provided at $s_t$ and $s_{t-1}$. Adding 0 or any of the potential deviations to the $course^t$ and $speed^t$ results to $(n+1)*(n+1)$ potential trajectory evolutions in $\textbf{T}_p^t$.

In the next section we discuss how we exploit the potential trajectory evolutions to estimate conflicts.

\subsection{Trajectory states, ATCO modes of reaction and resolution actions}
\label{sec:states-observations-modes-actions}

\begin{figure}[h]
\centering
\includegraphics[width=0.5 \linewidth]{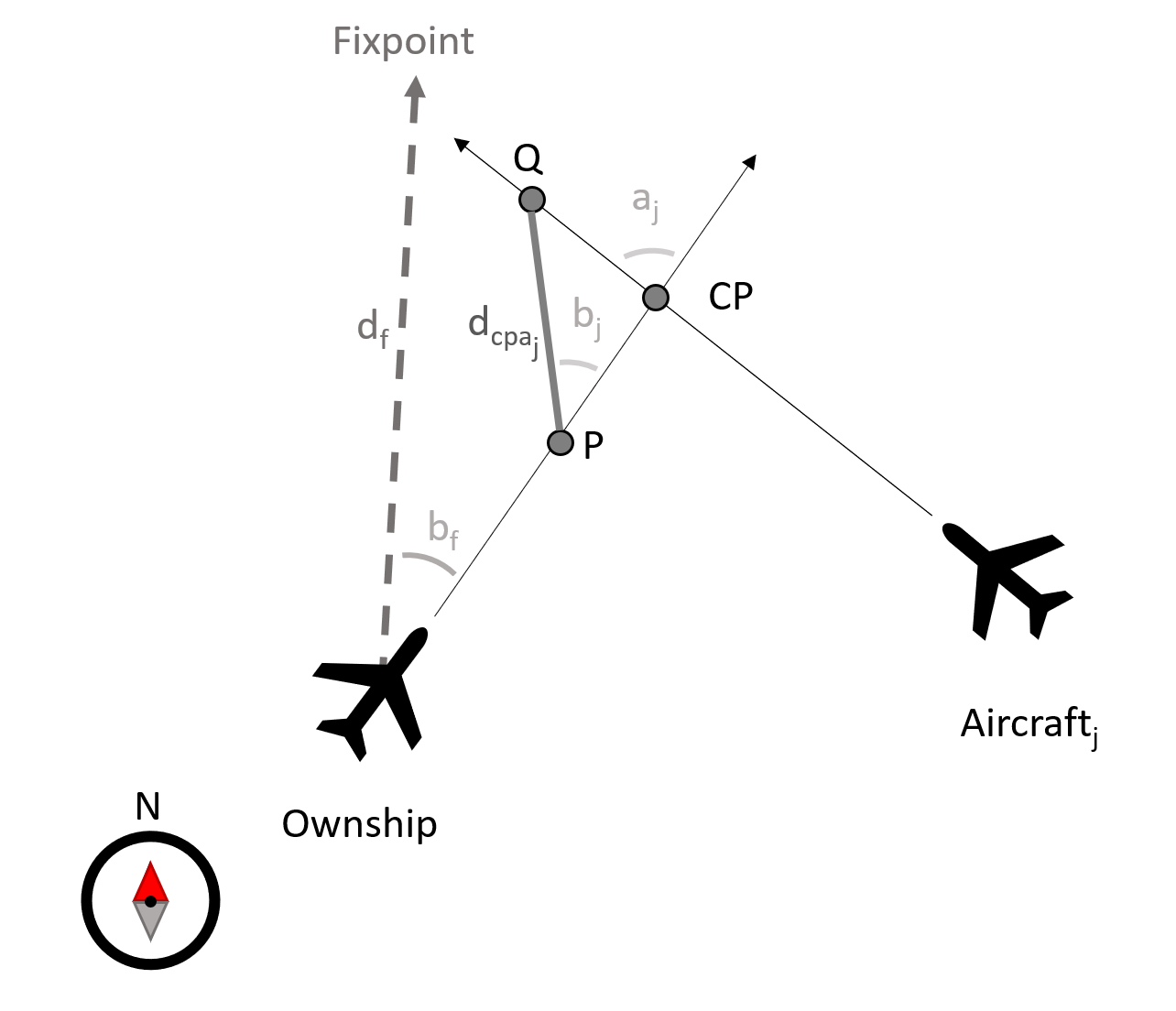}
\caption{ Features enriching a trajectory point w.r.t. the aircraft flying a conflicting trajectory $T_j$. }
\label{fig:EdgeFeatures}
\end{figure}

Motivated by the bibliography on CD\&R ( \cite{pham2019machine}, \cite{pham2019reinforcement}, \cite{ghosh2020deep}, \cite{dalmauair}), we enrich trajectory states with a vector $v$ of variables including, the magnitudes of the aircraft horizontal ($s_h$) and vertical ($s_v$) speed, as well as with the following features regarding any neighboring trajectory $T_j$:
\begin{center}
$\langle sin(b_f), cos(b_f), d_f , d_{h_{cpa_j}}, d_{v_{cpa_j}}, t_{cpa_j}, d_{cp_j}, t_{cp_j}, sin(a_j), cos(a_j), sin(b_j), cos(b_j) \rangle$
\end{center}
As Figure \ref{fig:EdgeFeatures} depicts, $b_f$ is the relative bearing w.r.t. a fixpoint, $d_f$ is the distance from that fixpoint, 
$d_{h_{cpa_j}}$, $d_{v_{cpa_j}}$ are the horizontal and vertical distance of the flights at the CPA and $t_{cpa_j}$ is the time of the ownship to CPA. $d_{cp_j}$ is the distance between the ownship and the neighbouring aircraft $j$, when the first of these is at the crossing point and $t_{cp_j}$ is the time until the first of the flights is at the crossing point. The intersection angle between the two trajectories is $a_j$, and $b_j$ is the relative bearing of the ownship w.r.t. the trajectory $T_j$ at the CPA.

To compute the CPA we follow the methodology presented in \cite{pham2019machine}.

 As we want our model to be agnostic to the specific spatio-temporal (longitude, latitude, timestamp) position, so as to generalize beyond specific areas and origin destination pairs, we train and test our models using an abstraction of states as follows:
\begin{center}
 $s=(h, s_h,s_v, f)$   
\end{center}
where, $f$ is a vector with features regarding all the neighboring trajectories.

 

     
    
 
As far as the actions are concerned, as described, we consider two levels of action abstractions: Modes of behaviour and types of resolution actions. 

The set of modes comprises three modes:

- $C_0$: No conflicts detected, and no resolution action is applied.

- $C_1$: At least one conflict is detected, and a resolution action is applied.

- $C_2$: At least one conflict is detected but no resolution action is applied.

The last mode indicates ATCO's tolerance to some conflicts, and allows delaying reactions after detecting conflicts and assessing the safety-criticality of a situation.

The set of types of resolution actions consist of categorical actions: 

-$A_0$:``No resolution action''

-$A_1$: ``Speed change resolution action''

-$A_2$: ``Direct to waypoint resolution action''

Although our main focus here is to predict the mode of the ATCO behaviour at specific times, the model is trained to also predict categorical conflict resolution actions as well as  continuous actions regarding the trajectory evolution. The set of continuous actions comprise the change in course $\Delta course$ of the trajectory, the change in the horizontal and vertical speeds $\Delta s_h$ and $\Delta s_v$ and also the time to the next point $\Delta t$. 

All these actions are important towards training models of timely ATCO reactions and specify the policy of the ATCO in conjunction to the policy of trajectory evolution at a fine level of detail. However, our focus in this paper is on predicting ATCO mode of behavior: Future work will exploit these models to predict ATCO resolution actions in conjunction to actions driving the evolution of the (conflicts-free) trajectory from state to state.

\subsection{Learning timely reactions}


As already described in Section \ref{sec:imitation}, towards predicting the ATCO' timely reactions we train a Variational Autoencoder (VAE) imitating the demonstrated ATCO policy in a supervised way. Modes of behaviour are decided by the encoder and exploited by the policy, which is represented by the decoder network and prescribes sequences of low-level actions. 

The encoder and decoder networks are trained by exploiting enriched trajectory points, the associated ATCO reaction modes and resolution actions. The errors regarding the predicted actions propagate backwards from the decoder. The encoder aims to minimize the categorical cross entropy loss between the distribution of modes in the dataset and the distribution predicted by the encoder.

To train the VAE for the continuous low-level actions we minimize the Mean Squared Error (MSE) and for the categorical actions the categorical cross entropy between the distribution of actions in the data set and the distribution of the decoder predictions. 

Formally, the loss function of VAE is as follows: 

$
\begin{aligned}
L_{VAE} (\pi,q)= & -\mathbb{E}_{(c^t\sim q,(a^t,s^t) \sim p_{data})} [log \pi_{\theta}(a^t|s^t,c^t )]\\ &-\mathbb{E}_{(c^t,c^{t-1},s^t) \sim p_{data} )} [log q_{\phi}(c^t | c^{(t-1)},s^t )]
\end{aligned}
$
\\

where $\pi_{\theta}$ is the decoder’s policy, $q_{\phi}$ is the encoder network, $a$, $s$ and $c$ are the actions, states and modes, respectively, $p_{data}$ denotes the data distribution and $t$ the timestep.

As modes are categorical variables we used the Gumbel-softmax trick \cite{jang2016categorical} to obtain samples from a categorical distribution.


Regarding the networks architecture, as $L_{VAE}$ specifies, the encoder network, given the mode at time point $t-1$ together with the state at time point $t$, predicts the mode at the current time point $t$. The decoder network takes as input the predicted mode and the state at the current time point $t$ and predicts the probabilities of low-level actions at time point $t$.

The encoder and decoder networks consist of two layers of 64 LSTM nodes each, with $tanh$ activation. Additionally, the encoder has a dense output layer with linear activation and number of nodes equal to the number of high level actions. Similarly, the decoder has a dense layer with linear activation and number of nodes equal to the number of categorical actions, and another dense output layer for the continuous actions, with linear activation and number of nodes equal to the number of continuous actions.
To minimize the loss function for both the Encoder and VAE we use the Adam optimizer.

\section{Experimental Evaluation}
\label{sec:experimental evaluation}
\subsection{Experimental Setting}

We evaluate the proposed method in two different types of settings w.r.t. the area of  responsibility (AoR) chosen: a) The sector-related and b) the sector-ignorant settings.

In the sector-related case the AoR $SA$ corresponds to a sector crossed by the trajectory of the ownship. Given that neighbouring flights are all flights in $SA$ following the constraints in $CR$ (according to the definition of neighboring flights), we set the horizontal distance threshold $D_{th}$ to infinity.

In the sector-ignorant case we simulate a setting close to the flight-centric one and consider a rectangular area covering the Iberian Peninsula. We segregate this area in cells of size 0.5 degrees longitude and latitude. We do so in order to create an index of the positions of flights in each cell at each time point. This allows fast access to the flights of each cell at each time point, making the identification of neighbouring flights more efficient in terms of computational time. For each trajectory point in this area we limit the neighbouring flights w.r.t. a focal trajectory to those with a distance threshold $D_{th}$ to 5 cells in the longitude (approx. 231 km) and latitude (approx. 308km) dimensions. This has as an effect that the area defined by the area covering  the Iberian Peninsula and $D_{th}$ specifies $SA$ and follows the movement of the ownship.

\begin{figure}[h]
\centering
\includegraphics[width=0.6\linewidth]{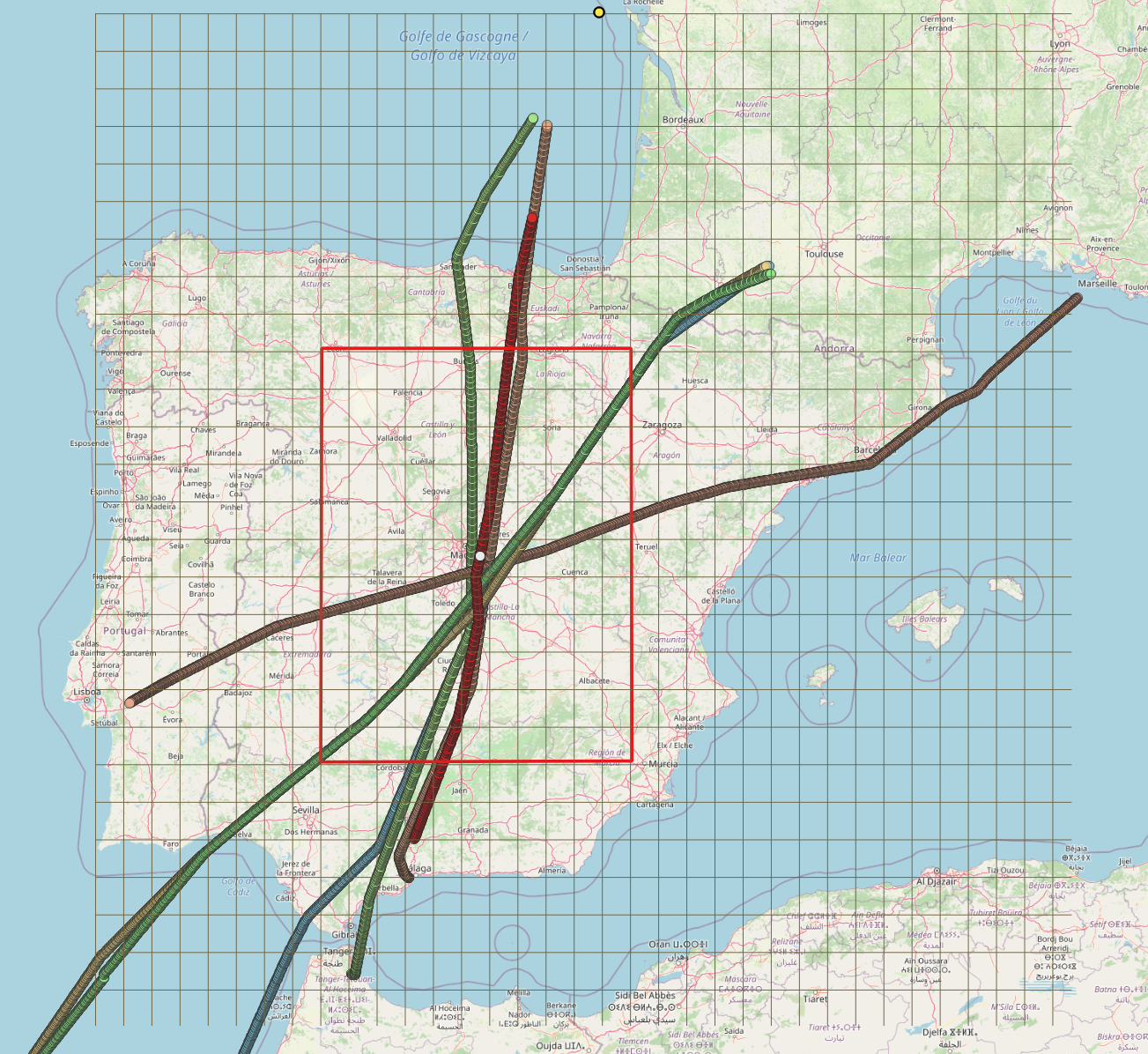}
\caption{The $SA$ area and the area defined by $D_{th}$ (red rectangular area) w.r.t. the ownship's position (white dot) in the sector-ignorant case.}
\label{fig:mbr}
\end{figure}

Figure \ref{fig:mbr} shows the $SA$ area considered in the sector-ignorant case (area covered by the grid), the focal trajectory of the ownship (red trajectory or dark gray in grayscale) and neighboring trajectories  in $Neigh(T_f,SA,t)$. The ownship's position at time $t$ is shown in white (middle point). The area defined by $D_{th}$ w.r.t. the ownship's position is depicted by the red rectangle. The yellow dot in the upper part of the grid is the fixpoint used and will be further discussed in the following sections.

\subsubsection{Data sets and Pre-processing}
\label{sec:Preprocessing}
As discussed in section \ref{sec:states-observations-modes-actions},  we annotate the trajectory points for all trajectories in $T_E$ using the  modes $C_0$ (``No conflicts detected, and no resolution action has been applied"), $C_1$ (``At least one conflict is detected, and a resolution action has been applied"), and $C_2$ (``At least one conflict is detected but no resolution action has been applied").

In so doing we have to deal with the following two problems:
\begin{enumerate}
	\item The dataset is imbalanced regarding the modes.  The typical case is to have one resolution action and one point with $C_1$ mode for a trajectory with 700 points (699 points corresponding to modes $C_0$ and $C_2$).
	\item Following a data-driven approach and exploiting data with flown (thus conflict-free) trajectories, although we have defined the methodology presented in Section \ref{sec:simulating uncertainty} to detect conflicts that may exist prior to issuing a conflict resolution action recorded in the datasets, there are cases where there is an ATCO resolution action for a trajectory but no conflicts are detected.
\end{enumerate}

We cope with the first problem by augmenting data: The trajectory points in a time window of 250 seconds before the resolution action are annotated with $C_1$, except the points at which no conflicts are detected which are filtered out. This, somehow incorporates the uncertainty of ATCO’s decision about the time to issue a resolution action. 
We call the point at which the resolution action was issued the \textit{actual} Resolution Action Trajectory Point (RATP) and the annotated points due to data augmentation, before the \textit{actual} RATP, are called \textit{annotated} RATPs.

Next, we apply subsampling to trajectory points with modes $C_0$ and $C_2$, keeping one trajectory point with any of these modes every $step$ trajectory points. We experimented with different step sizes in order to balance the prior distribution between modes.
Table \ref{table:prior} reports the prior distribution of modes $C_0$, $C_1$ and $C_2$ for different step sizes considering the dataset of the sector-ignorant experimental setting. Given these distributions we decided to set $step$ equal to 6 subsequent trajectory points (i.e. 30 seconds).

\begin{table}
\centering
\footnotesize
\begin{tabular}{|c|c|c|c|}
\hline
$step$ &	$C_0$ & $C_1$ & $C_2$ \\
\hline
1 & 0.63933797 & 0.05469366 & 0.30596837 \\
2 & 0.61360892 & 0.10181661 & 0.28457447 \\
3 & 0.59235067 & 0.14231008 & 0.26533926 \\
4 & 0.57311712 & 0.17803213 & 0.24885076 \\
5 & 0.55696083 & 0.20905254 & 0.23398663 \\
6 & 0.54101562 & 0.23763021 & 0.22135417 \\
\hline
\end{tabular}
\caption{ Prior distribution of modes ($C_0$, $C_1$, $C_2$) for different subsampling $step$ values 
computed on the dataset for the sector-ignorant experimental setting.}    
\label{table:prior}
\normalsize
\end{table}

Regarding the second problem, we filter out trajectories where there is an associated ATCO resolution action in the dataset but no conflicts are detected by our methodology in a time window of $window\_duration$ seconds before the actual RATP. The trajectory point (if any) in the specified time window at which at least one conflict is detected and is temporally closest to the point where the ATCO resolution action is indicated, as the \textit{actual} RATP. We set $window\_duration$=70s. This choice follows consistently the evaluation methodology that is described in Section \ref{sec:evaluation}.

The fixpoint of a flight in the sector-related setting is the exit point from the sector. In the sector-ignorant setting is the point at which the  edge towards the destination airport of the $SA$ box crosses the  line connecting the origin and the destination airports. 

The dataset contains trajectories between 5 different origin-destination pairs, all from 2017: Malaga (LEMG) - Gatwick (EGKK), Malaga (LEMG) - Amsterdam (EHAM), Lisbon (LPPT)- Paris(LFPO), Zurich (LSZH) - Lisbon (LPPT) and Geneva (LSGG) - Lisbon (LPPT). We study only ATCO resolution actions issued at the en-route phase of operations and filter out the climb and descent parts of the trajectories. In addition, we consider only trajectories that have at least one ATCO resolution action and an associated actual RATP.  This results to 255 enriched trajectories corresponding to 344 resolution actions for the sector relevant case and 668 trajectories corresponding to 791 resolution actions for the sector-ignorant case. It must be noted here that the available ATCO events dataset covers the Spanish airspace and thus we consider the points of the trajectories that are in this airspace. However, the proposed method is generic, goes beyond sectors, as we show in the sector-ignorant case, and can be applied in any airspace.



\subsubsection{Evaluation methodology}
\label{sec:evaluation}

To evaluate the accuracy of predictions made by the proposed models we define weighted versions of precision and recall: We do so, in order to provide the flexibility needed for the predicted RATPs, compared to the actual and annotated RATPs. This flexibility is necessary, as  ATCO' timely reactions may differ even in the same situation, if this situation occurs at different times and/or for different ATCO. Weighted versions of precision and recall  penalize predicted RATPs based on their temporal distance to the actual or annotated RATPs.  for this purpose  we define a score function based on temporal distance between RATPs using a Gaussian distribution with $std=$5, as justified below. 

Formally the score function is as follows:

\begin{center}
    $score(x)= \frac{ \frac{1}{5\sqrt{2\pi}} e^{-\frac{1}{2} ({\frac{x}{5n})^2 }}}{\frac{1}{5\sqrt{2\pi}} e^{-\frac{1}{2} ({\frac{0}{5n})^2 }} } = e^{-\frac{1}{2} ({\frac{x}{5n})^2 }}$
\end{center}

The parameter $n$ is a factor translating the temporal distance in the x-axis  to a number of n-sec-intervals. We set $n$ equal to 5, given that decisions are made using the temporal granularity of 5 seconds. By using different values for the standard deviation and $n$ we can tune this score function, which may be considered to estimate the probability that a reaction happens with a temporal difference $x$ compared to the ATCO reaction recorded in the data. 
\begin{figure}[h]

\includegraphics[width=\linewidth]{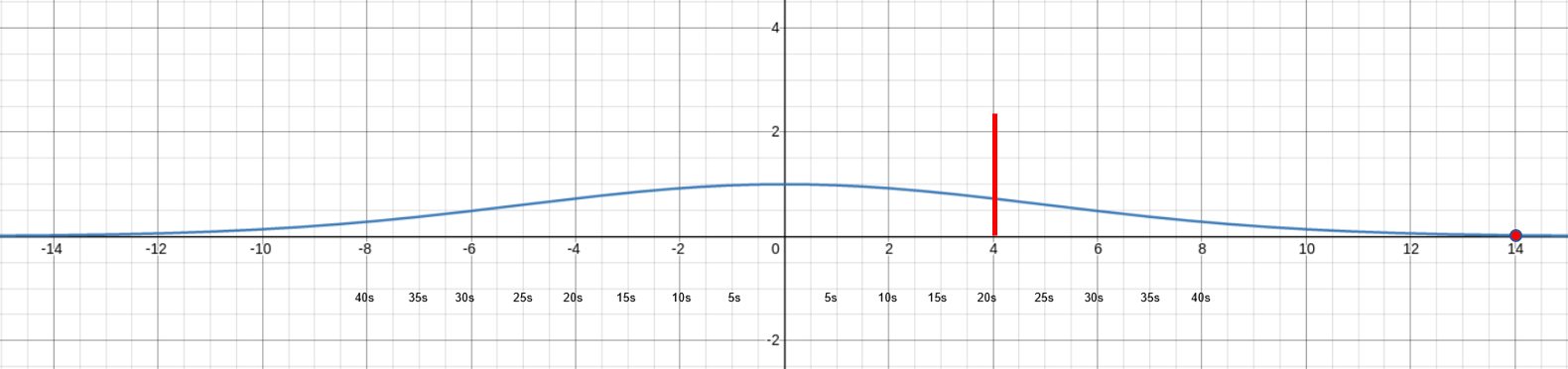}
\caption{Score function}
\label{fig:eval_func}
\end{figure}

This function is depicted in Figure \ref{fig:eval_func}, taking values between 0 and 1.  Setting $n$ to 5 we sync the temporal distance to the change of slope when $x$=20 seconds at $x/5$=4 (indicated by a red line in Figure \ref{fig:eval_func}): Thus, the score is reduced more drastically when the temporal distance between the predicted and the actual reactions is greater than 20 seconds. The parameter $n$ also controls the time window out of which the score is approximately 0. For $n$=5 this is set to approx. 70s (shown in Figure \ref{fig:eval_func} with the red dot at the right). This is also the value used for the  $window\_duration$ used when setting the actual RATP during preprocessing, as mentioned in Section \ref{sec:Preprocessing}.

Next, we discuss how the score function is used  to calculate precision and recall in experiments,  weighting true/false positives/negatives per type of reaction.  The rationale for the weighting scheme is that we want to penalize false predictions according to their temporal distance from the closest in time actual or annotated RATP. For example, let us consider a case where the model reacts to a conflict  5 seconds later or earlier than the reaction recorded in the dataset. This, due to the small temporal distance between the predicted and the actual RATPs, will be penalized lightly, assuming that both decisions were driven by the same contextual features. On the other hand, a large temporal distance may be due either to different contexts in which reactions occur, or to the lack of model's capability to react at the right time. In either case, the predicted reaction differs considerably from the one provided in the data set. 

To explain the evaluation scheme, let us consider the (high and low level) reactions in their most general form. We distinguish the following two generic classes of ATCO reactions: 

- $G_0$: Not Assigning Resolution Action

- $G_1$: Assigning Resolution Action
    
Class $G_0$ includes modes $C_0$ and $C_2$ and resolution action $A_0$, whereas class $G_1$ includes mode $C_1$ and resolution actions $A_1$,$A_2$. Thus both $G_0$ and $G_1$ cases are specialized by  subclasses of modes and low-level types of ATCO resolution actions. Subsequently, we denote $G_{0_j}, j=1,..,k$ and $G_{1_j}, j=1,..,l$ the different subclasses of each class $G_0$ and $G_1$, respectively. The methodology can be generalized beyond the classes considered here. 

Next we describe in detail how false and true positives/negatives for $G_0$ and $G_1$ are weighted when computing precision and recall. Although cases are presented starting from the most general classes, it must be noted that the weighted scheme applies to either modes or ATCO resolution actions.

1. \textbf{False positives/negatives of a subclass} $G_{0_j}$ 

    1(a) \textit{False positives}: In this case the model falsely predicts a subclass $G_{0_j}$ of $G_0$ while the dataset indicates either $G_1$ ``Assigning Resolution Action" or a subclass $G_{0_k}$ of $G_0$ with $j \neq k$.
    
        1(a)i. \label{it:G0fpa} Let us consider the case where the dataset indicates $G_1$ ``Assigning Resolution Action". We penalize the prediction according to its distance from the closest actual RATP. Our rational here is that if the prediction corresponds to an annotated point far from the actual RATP then we want to penalize it lightly as a false positive. On the other hand, if the model does not predict a resolution action close to the actual RATP then we want to penalize it heavily. To do so, given the time point of the prediction $t_p$ and the time point of the closest actual RATP $t_a$, we calculate the temporal distance $x=|t_p-t_a|$ and the weight $w_{fp}=fscore(x)$, considering that the prediction is false positive with weight  $w_{fp}$ and  true positive with weight $(1-w_{fp})$. This means that $G_0$ false (true)  positives are weighted by $w_{fp_i}$ (resp., $(1-w_{fp_i})$) (resulting to $\sum_{i=1}^{\#FP}w_{fp_i}$, resp. to $\sum_{i=1}^{\#FP}(1-w_{fp_i})$).
        
         1(a)ii. \label{it:G0fpb} Regarding the second case, the model predicts a subclass $G_{0_j}$ of $G_0$ while the dataset indicates another subclass $G_{0_k}$ of $G_0$ with $j \neq k$. As ATCO resolution actions in $G_0$ can be only of type $A_0$, this case may occur only when considering modes. For example, let us consider the case where the mode indicated in the dataset is $C_2$: ``At least one conflict is detected but no resolution action has been applied" and the model indicates $C_0$: ``No conflicts detected, and no resolution action has been applied". This is clearly a false prediction that is penalized heavily by assigning a weight $w=1$.

    1(b) \textit{False negatives}:    
    In this case the model falsely predicts either $G_1$ ``Assigning Resolution Action" or a subclass of $G_0$, $G_{0_k}$, $k \neq j$, while the dataset indicates $G_{0_j}$.
    
     1(b)i. Let us consider the case where the model predicts $G_1$:``Assigning Resolution Action" instead of $G_{0_j}$:``Not Assigning Resolution Action". If this falsely predicted point is close to either an annotated or an actual RATP we want to tolerate the error penalizing lightly the prediction. To do so, given the time point of the prediction $t_p$ and the time point of the closest RATP (either actual or annotated) $t_a$, we calculate the temporal distance $x=|t_p-t_a|$ and the weight $w_{fn}=1-fscore(x)$, considering that the prediction is false negative with weight  $w_{fn}$ and  true negative with weight $(1-w_{fn})$. As is also evident by the formulas for $w$, $w_{fn}=1-fscore(x)$ and $w_{fp}=fscore(x)$, this case is the complementary of case 1(a)i.
       
     1(b)ii. Considering the case where the dataset indicates $G_{0_j}$ but the model falsely predicts another subclass of $G_0$, $G_{0_k}$ where $k \neq j$, similarly to the case  1(a)ii. 
     , we assign a weight $w=1$ considering that the prediction is a false negative with weight of 1.

2. \textbf{False positives/negatives of a subclass} $G_{1_j}$ 

    2(a) \textit{False positives}: In this case the model falsely predicts a subclass $G_{1_j}$ of $G_1$ ``Assigning Resolution Action" while the dataset indicates either $G_0$ ``Not Assigning Resolution Action" or a subclass $G_{1_k}$ of $G_1$ with $j \neq k$.
    
    2(a)i. \label{it:G1fpa} Let us consider first the case where the dataset indicates $G_0$ ``Not Assigning Resolution Action" at a point, but the model does issue a resolution action. If the false prediction is close to either an annotated or an actual RATP,  the prediction is penalized lightly. To do so, given the time point of the prediction $t_p$ and the time point of the closest RATP (either actual or annotated) $t_a$, we calculate the temporal distance $x=|t_p-t_a|$ and the weight $w_{fp}=1-fscore(x)$ considering that the prediction is false positive with weight  $w_{fp}$ and  true positive with weight $(1-w_{fp})$.
    
    2(a)ii. \label{it:G1fpb} Considering the second case,  as there is a single mode $C_1$ in class $G_1$, this case occurs only when considering ATCO resolution action subclasses. In such cases we do not tolerate errors as the model predicts a wrong resolution action according to the one demonstrated in the dataset. Thus we assign a weight $w=1$ considering that the prediction is a false positive with weight of 1.

    2(b) \textit{False negatives}:    
    In this case the dataset indicates $G_{1_j}$ but the model falsely predicts either $G_0$ ``Not Assigning Resolution Action" or a subclass of $G_1$, $G_{1_k}$ where $k \neq j$.
    
     2(b)i. Let us consider the case where the model predicts $G_0$ "Not Assigning Resolution Action" instead of $G_{1_j}$:"Assigning Resolution Action". The prediction is penalized according to the distance of the prediction from the closest actual RATP. Our rational here, as in case  1(a)i.
     , is as follows: if the prediction corresponds to an annotated point far from the actual RATP then we want to penalize it lightly as a false negative. On the other hand, if the model does not predict a resolution action close to the actual RATP then we want to penalize it heavily. To do so, given the time point of the prediction $t_p$ and the time point of the closest actual RATP $t_a$ we calculate the temporal distance $x=|t_p-t_a|$ and the weight $w_{fn}=fscore(x)$ considering that the prediction is false negative with weight  $w_{fn}$ and  true negative with weight $(1-w_{fn})$. Again this means that false (true)  negatives are weighted by $w_{fn_i}$ (resp., $(1-w_{fn_i})$). 
        
        2(b)ii. Considering the case where the dataset indicates $G_{1_j}$ but the model falsely predicts a subclass 
        $G_{1_k}$, where $k \neq j$, then, as specified in case 2(a)ii.
        , we assign a weight $w=1$ considering that the prediction is a  false negative with weight 1.

3. \textbf{True positives of a subclass} $G_{i_j}$: True positives are those cases where the model correctly predicts a subclass $G_{i_j}$ (assigned a score equal to 1). Additionally, given that false positives  are assessed with weight $w_{fp}$, then for the corresponding cases  we may have a true positive with weight  $(1-w_{fp})$. Thus, true positives are calculated using the following formula: $\sum_{i=1}^{\#TP}1 + \sum_{i=1}^{\#FP}(1-w_{fp_i})$

4. \textbf{True negatives of a subclass} $G_{i_j}$: As true negatives we consider those cases where the model correctly does not predict a subclass $G_{i_j}$ and it is scored with weight 1. In addition, given that false negatives predicted are assessed with $w_{fn}$, then with weight $(1-w_{fn})$ we may have a true negative for the corresponding cases. Thus, true negatives are calculated using the following formula: $\sum_{i=1}^{\#TN}1 + \sum_{i=1}^{\#FN}(1-w_{fn_i})$

	Having considered the different cases we formally define the weighted version of precision, recall and f1-score, namely WP, WR and WF1 respectively, as follows:

\begin{equation}
\label{eq:wp}
\begin{aligned}
    WP &=\frac{TP}{TP+FP} \\
    &= \frac{\sum_{i=1}^{\#TP}1 + \sum_{i=1}^{\#FP}(1-w_{fp_i})}{[\sum_{i=1}^{\#TP}1 + \sum_{i=1}^{\#FP}(1-w_{fp_i})] + \sum_{i=1}^{\#FP}w_{fp_i}} \\
    &= \frac{\sum_{i=1}^{\#TP}1 + \sum_{i=1}^{\#FP}(1-w_{fp_i})}{\sum_{i=1}^{\#TP}1 + \sum_{i=1}^{\#FP}1}
\end{aligned}
\end{equation}    
\begin{equation}
\label{eq:wr}
\begin{aligned}
    WR &= \frac{TP}{TP+FN} \\
    &= \frac{\sum_{i=1}^{\#TP}1 + \sum_{i=1}^{\#FP}(1-w_{fp_i})}{[\sum_{i=1}^{\#TP}1 + \sum_{i=1}^{\#FP}(1-w_{fp_i})] + \sum_{i=1}^{\#FN}w_{fn_i}} \\
\end{aligned}
\end{equation}
\begin{equation}
\label{eq:wf1}
    WF1=2*\frac{WP*WR}{WP+WR}
\end{equation}
	
In these formulae, \#TP is the number of true positives, \#FP is the number of false positives and \#FN is the number of false negatives, 
It must be noticed that when $w_{fp_i}$ and $w_{fn_i}$ is equal to 1 WP, WR and WF1 revert to the standard precision, recall and F1 measures.
 



\subsection{Experimental Results}


In the following we report the results achieved by the VAE model and we also report the results achieved by training only the encoder network (baseline). This shows the difference in performance between the two methods, caused by the decoder's error backwards propagation.
To evaluate the VAE and the baseline method  we perform 10 experiments with two times repeated 5-fold cross validation, training the models for 1000 epochs at each experiment.
We report the 95\% confidence interval (CI) of the non-weighted precision, recall and f1-score, in conjunction to the weighted versions of these measures.

\begin{table}[]
\tiny
    \centering
    \begin{tabular}{c|c|c|c|c}
         model & modes non-weighted & modes weighted & actions non-weighted & actions weighted  \\
         \hline \hline
         VAE & 
             \begin{tabular}{c}
                precision \\ \hline
                $C_0: 1.000\pm0.000$ \\
                $C_1: 0.976 \pm 0.006$\\
                $C_2: 0.934 \pm 0.012$\\ \hline
                recall \\ \hline
                $C_0: 1.000 \pm 0.000$\\
                $C_1: 0.936 \pm 0.014$\\
                $C_2: 0.976 \pm 0.006$\\ \hline
                f1-score \\ \hline
                $C_0: 1.000 \pm 0.000$\\
                $C_1: 0.956 \pm 0.008$\\
                $C_2: 0.954 \pm 0.008$
             \end{tabular}
             &              
             \begin{tabular}{c}
                precision \\ \hline
                $C_0: 1.000 \pm 0.000$ \\
                $C_1: 0.982 \pm 0.005$\\
                $C_2: 0.990 \pm 0.000$\\ \hline
                recall \\ \hline
                $C_0: 1.000 \pm 0.000$\\
                $C_1: 0.989 \pm 0.002$\\
                $C_2: 0.983 \pm 0.005$\\ \hline
                f1-score \\ \hline
                $C_0: 1.000 \pm 0.000$\\
                $C_1: 0.985 \pm 0.004$\\
                $C_2: 0.986 \pm 0.004$
             \end{tabular} 
             & 
            \begin{tabular}{c}
                precision\\ \hline
                $A_0: 0.975 \pm 0.004$\\
                $A_1: 0.635 \pm 0.028$\\
                $A_2: 0.646 \pm 0.035$\\ \hline
                recall\\ \hline
                $A_0: 0.990 \pm 0.000$\\
                $A_1: 0.549 \pm 0.026$\\
                $A_2: 0.670 \pm 0.022$\\ \hline
                f1-score\\ \hline
                $A_0: 0.985 \pm 0.004$\\
                $A_1: 0.588 \pm 0.017$\\
                $A_2: 0.656 \pm 0.021$\\
             \end{tabular} 
             &
             \begin{tabular}{c}
                precision\\ \hline
                $A_0: 0.998 \pm 0.003$\\
                $A_1: 0.640 \pm 0.026$\\
                $A_2: 0.653 \pm 0.035$\\ \hline
                recall\\ \hline
                $A_0: 0.993 \pm 0.003$\\
                $A_1: 0.589 \pm 0.028$\\
                $A_2: 0.709 \pm 0.021$\\ \hline
                f1-score\\ \hline
                $A_0: 0.993 \pm 0.003$\\
                $A_1: 0.610 \pm 0.017$\\
                $A_2: 0.679 \pm 0.023$\\
             \end{tabular} \\
             \hline \hline
         Enc 
             &
            \begin{tabular}{c}
                precision\\ \hline
                $C_0: 1.000 \pm 0.000$\\
                $C_1: 0.950 \pm 0.010$\\
                $C_2: 0.870 \pm 0.038$\\ \hline
                recall\\ \hline
                $C_0: 1.000 \pm 0.000$\\
                $C_1: 0.863 \pm 0.053$\\
                $C_2: 0.951 \pm 0.011$\\ \hline
                f1-score\\ \hline
                $C_0: 1.000 \pm 0.000$\\
                $C_1: 0.904 \pm 0.032$\\
                $C_2: 0.909 \pm 0.020$\\
             \end{tabular}
             &
            \begin{tabular}{c}
                precision\\ \hline
                $C_0: 1.000 \pm 0.000$\\
                $C_1: 0.959 \pm 0.009$\\
                $C_2: 0.975 \pm 0.009$\\ \hline
                recall\\ \hline
                $C_0: 1.000 \pm 0.000$\\
                $C_1: 0.969 \pm 0.017$\\
                $C_2: 0.961 \pm 0.011$\\ \hline
                f1-score\\ \hline
                $C_0: 1.000 \pm 0.000$\\
                $C_1: 0.964 \pm 0.009$\\
                $C_2: 0.967 \pm 0.006$\\
             \end{tabular}
             & - & - \\
    \end{tabular}
    \caption{ Experimental Results of the sector-ignorant case achieved by the VAE and the Encoder (Enc). Columns report the 95\% confidence interval of precision, recall and f1-score w.r.t. the modes and the resolution actions of ATCO, for the non-weighted and weighted measures.}
    \label{tab:SI_results}
\end{table}

Table \ref{tab:SI_results} reports the 95\% confidence interval of the precision, recall and f1-score, achieved by the VAE and the Encoder (Enc) for the ATCO modes of behavior and the resolution actions, for the sector-ignorant case. Columns ``modes non-weighted"/``actions non-weighted" and ``modes weighted" / ``actions weighted" report respectively on the non-weighted and the weighted versions of the measures for modes and resolution actions. As the encoder does not predict resolution actions, corresponding columns in the second row are empty.

Regarding the modes of ATCO reaction, results show that both the VAE and the Encoder networks achieve over 0.9 f1-score on all modes, for  the non-weighted and the weighted measures, with VAE achieving the best results with a weighted f1-score greater or equal to $0.985\pm0.004$ on all modes. Also the VAE  outperforms the encoder  on all measures, weighted or not, although the Encoder is really competitive. The largest differences between the models are observed w.r.t. the precision of mode $C_2$ and the recall of mode $C_1$ for the non-weighted measures. For mode $C_2$ VAE achieves a precision of $0.934\pm0.012$ whereas the Encoder achieves a precision of $0.870\pm0.038$. For mode $C_1$ the VAE and the Encoder achieve a recall of $0.936\pm0.014$ and $0.863\pm0.053$, respectively. This shows that there are cases where the Encoder should assign a resolution action but it fails to do so, as it predicts mode $C_2$. For the VAE, such cases are rather rare.

An interesting observation is that the non-weighted precision of mode $C_1$ is higher of that of mode $C_2$, while the non-weighted recall of mode $C_1$ is lower than that of mode $C_2$. Regarding the weighted measures the situation is quite the opposite, as the weighted precision of mode $C_1$ is lower of that of mode $C_2$, while the weighted recall of mode $C_1$ is higher than that of mode $C_2$. Regarding the differences between the weighted and non-weighted measures these can be explained as follows (we further delve into this in subsequent paragraphs): According to the non-weighted measures that penalize all errors equally, there are cases where the model should react by predicting mode $C_1$ at a specific trajectory point, but it does not, as it instead predicts mode $C_2$. On the other hand, when considering the weighted measures, we can conclude that the model predicts mode $C_1$ \textit{near} the RATPs, in  points that according to the dataset are annotated as $C_2$. These are mostly points that succeed the actual RATPs and precede the start of the manoeuvre that implements the resolution action. Note that manoeuvres implementing the resolution actions do not begin instantly after the ATCO reaction, as pilots need some time to react to the ATCO's instruction. Points succeeding the actual RATP and preceding the start of the manoeuvre have features that are close to the features of the corresponding actual RATPs, so they are penalized lightly.

\begin{table}[]
\centering
\tiny
    \begin{tabular}{c|c}
        Expert & Predicted\\ \hline
        \includegraphics[width=0.3\linewidth]{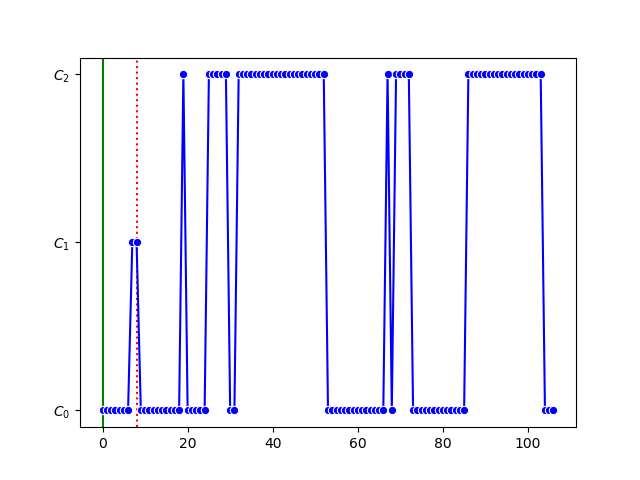} &
        \includegraphics[width=0.3\linewidth]{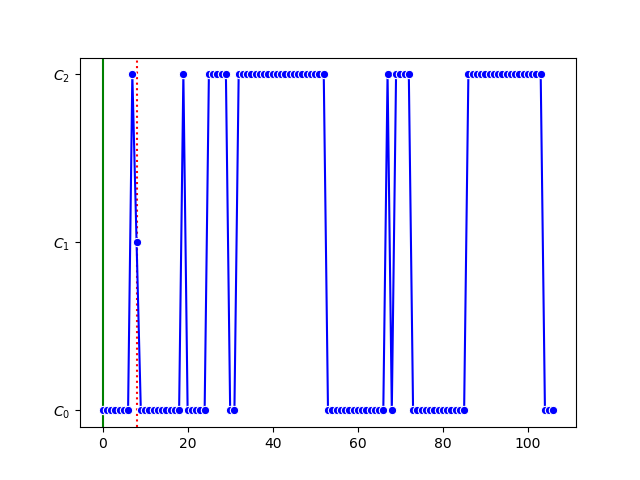}\\
        \hline
        \includegraphics[width=0.3\linewidth]{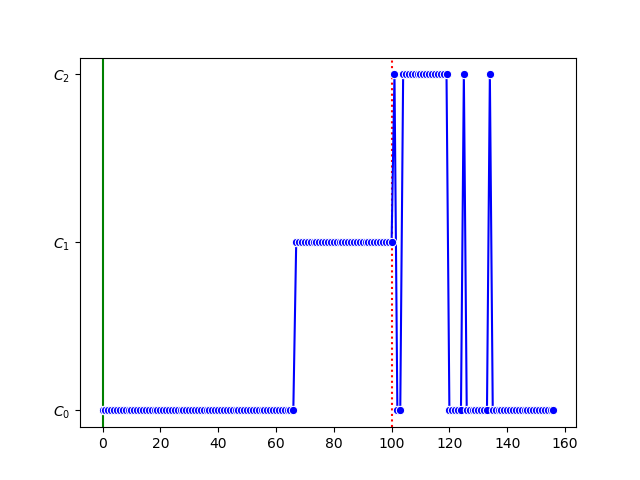} &
        \includegraphics[width=0.3\linewidth]{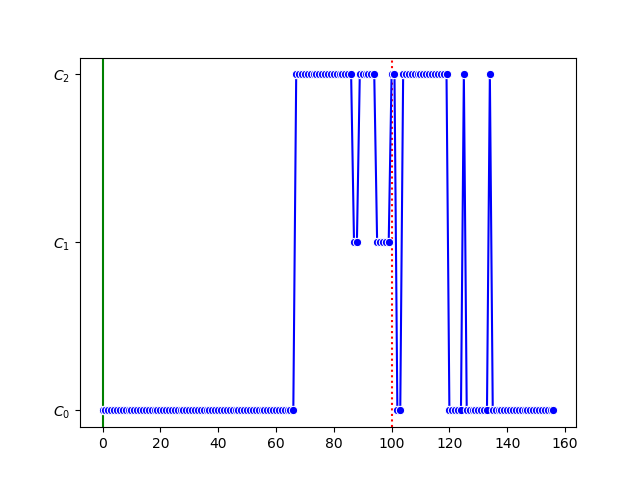}\\
        \hline
        \includegraphics[width=0.3\linewidth]{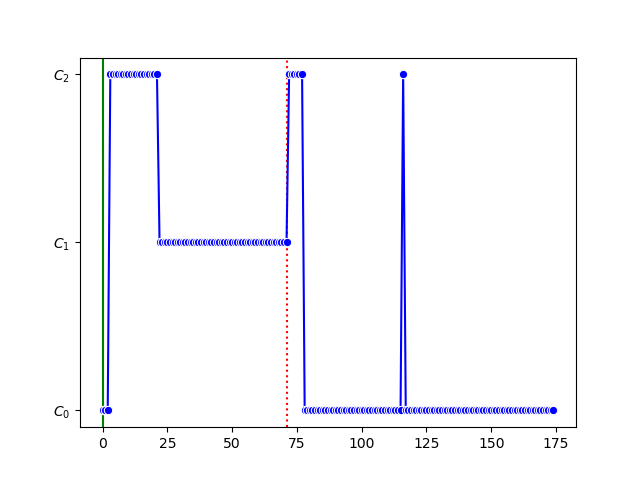} &
        \includegraphics[width=0.3\linewidth]{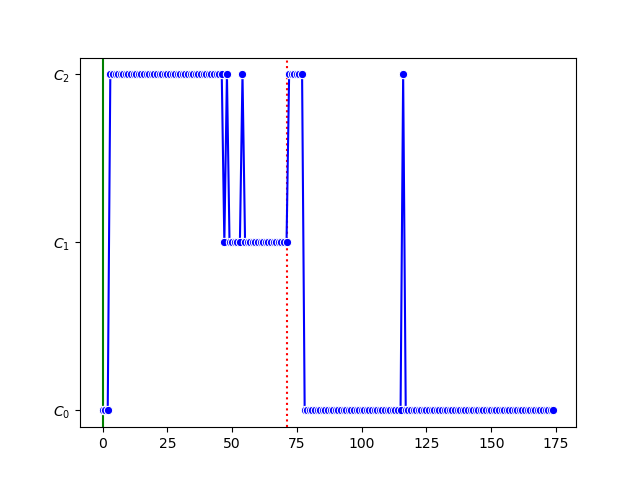}\\
        \hline
        \includegraphics[width=0.3\linewidth]{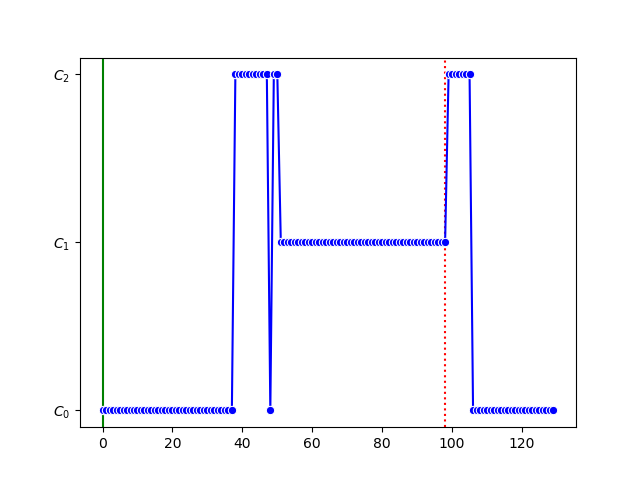} &
        \includegraphics[width=0.3\linewidth]{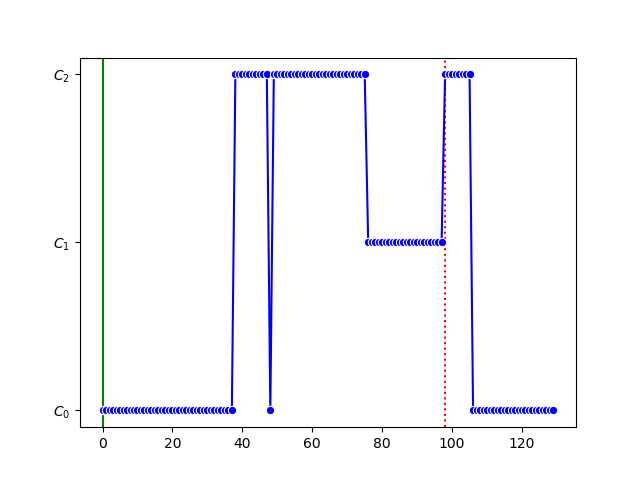}\\
        \hline
        \includegraphics[width=0.3\linewidth]{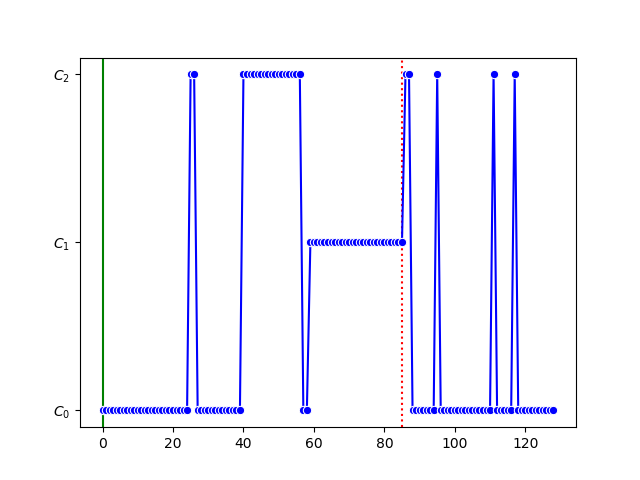} &
        \includegraphics[width=0.3\linewidth]{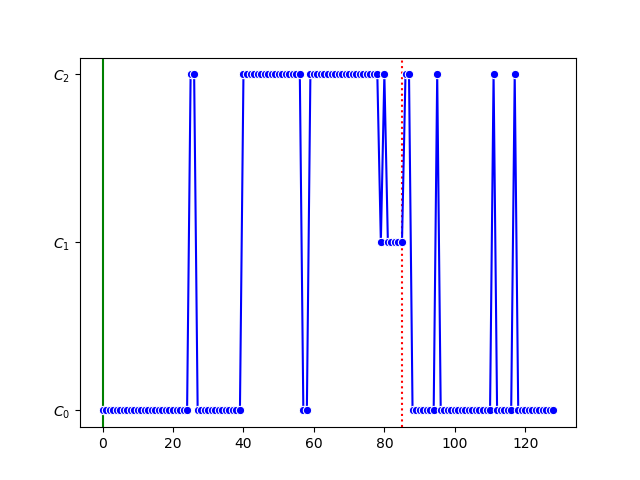}\\ 
    \end{tabular}
    \caption{The 5 trajectories with the highest difference between the weighted and non-weighted f1-score. Column ``Predicted" shows the modes predicted by the VAE model, whereas column ``Expert" shows the modes reported in the dataset. X-axis:  the sequence number of the states of all trajectories. Y-axis: the modes. Blue dots denote the mode at each point. Green solid vertical lines show the start of each trajectory, while red dashed ones show the actual RATP.}
    \label{tab:HighRecallDiff}
\end{table}


Results for the prediction of ATCO resolution actions are not so impressive as those achieved on the prediction of modes. The non-weighted f1-score for the $A_1$ and $A_2$ resolution actions is $A_1: 0.588\pm0.017$ and $A_2: 0.656\pm0.021$. The weighted f1-score is $A_1: 0.610\pm0.017$ and $A_2: 0.679\pm0.023$. The prediction of ATCO resolution actions will be further explored in future work.

Observing the weighted and non-weighted measures w.r.t. the modes of the ATCO reactions we see that the weighted measures are higher than the non-weighted. To better understand the difference between the non-weighted and the weighted measures, Table 3 shows 5 trajectories with the highest difference between the weighted and non-weighted f1-score  for the sector-ignorant case. Column ``Predicted" shows the modes predicted by the VAE model, whereas column ``Expert" shows the modes reported in the dataset. In the x-axis we report the sequence number of each state of all 5 trajectories and in the y-axis the modes. Blue dots denote the mode, either predicted or reported in the dataset, at each point. Green solid vertical lines denote the start of each trajectory, while red dashed ones denote the actual RATP. We observe that in all 5 trajectories the model predicts mode $C_1$ near the actual RATPs, which agrees with the annotation of the dataset. Also, the model predicts mode $C_2$ instead of mode $C_1$ at points which are further from the actual RATPs (in most cases), and such errors will be penalized lightly by our weighted measures. 

\begin{table}[]
\tiny
    \begin{tabular}{c|c|c|c|c}
         model & modes non-weighted & modes weighted & actions non-weighted & actions weighted  \\ \hline\hline
         VAE 
            &
            \begin{tabular}{c}
                precision\\ \hline
                $C_0: 1.000 \pm 0.000$\\
                $C_1: 0.919 \pm 0.029$\\
                $C_2: 0.791 \pm 0.048$\\  \hline
                recall\\ \hline
                $C_0: 1.000 \pm 0.000$\\
                $C_1: 0.835 \pm 0.045$\\
                $C_2: 0.893 \pm 0.036$\\ \hline
                f1-score\\ \hline
                $C_0: 1.000 \pm 0.000$\\
                $C_1: 0.873 \pm 0.025$\\
                $C_2: 0.835 \pm 0.025$\\
             \end{tabular}
             &
            \begin{tabular}{c}
                precision\\ \hline
                $C_0: 1.000 \pm 0.000$\\
                $C_1: 0.929 \pm 0.028$\\
                $C_2: 0.960 \pm 0.011$\\ \hline
                recall\\ \hline
                $C_0: 1.000 \pm 0.000$\\
                $C_1: 0.965 \pm 0.012$\\
                $C_2: 0.919 \pm 0.033$\\ \hline
                f1-score\\\hline
                $C_0: 1.000 \pm 0.000$\\
                $C_1: 0.945 \pm 0.015$\\
                $C_2: 0.940 \pm 0.014$\\
             \end{tabular}             
             & 
             \begin{tabular}{c}
                precision\\ \hline
                $A_0: 0.952 \pm 0.012$\\ \hline
                $A_1: 0.604 \pm 0.087$\\
                $A_2: 0.439 \pm 0.099$\\ \hline
                recall\\ \hline
                $A_0: 0.981 \pm 0.007$\\
                $A_1: 0.566 \pm 0.070$\\
                $A_2: 0.362 \pm 0.076$\\ \hline
                f1-score\\ \hline
                $A_0: 0.966 \pm 0.005$\\
                $A_1: 0.569 \pm 0.039$\\
                $A_2: 0.384 \pm 0.075$\\
             \end{tabular}
             &
            \begin{tabular}{c}
                precision\\ \hline
                $A_0: 0.993 \pm 0.003$\\
                $A_1: 0.611 \pm 0.088$\\
                $A_2: 0.446 \pm 0.100$\\ \hline
                recall\\ \hline
                $A_0: 0.983 \pm 0.006$\\
                $A_1: 0.661 \pm 0.068$\\
                $A_2: 0.428 \pm 0.093$\\ \hline
                f1-score\\ \hline
                $A_0: 0.986 \pm 0.004$\\
                $A_1: 0.620 \pm 0.036$\\
                $A_2: 0.419 \pm 0.076$\\
             \end{tabular}\\
         \hline \hline
         Enc 
             &
             \begin{tabular}{c}
                precision\\ \hline
                $C_0: 1.000 \pm 0.000$\\
                $C_1: 0.863 \pm 0.020$\\
                $C_2: 0.740 \pm 0.037$\\ \hline
                recall\\ \hline
                $C_0: 1.000 \pm 0.000$\\
                $C_1: 0.805 \pm 0.031$\\
                $C_2: 0.809 \pm 0.024$\\ \hline
                f1-score\\ \hline
                $C_0: 1.000 \pm 0.000$\\
                $C_1: 0.833 \pm 0.022$\\
                $C_2: 0.774 \pm 0.024$\\
             \end{tabular}
             &
             \begin{tabular}{c}
                precision\\ \hline
                $C_0: 1.000 \pm 0.000$\\
                $C_1: 0.874 \pm 0.021$\\
                $C_2: 0.950 \pm 0.009$\\ \hline
                recall\\ \hline
                $C_0: 1.000 \pm 0.000$\\
                $C_1: 0.955 \pm 0.011$\\
                $C_2: 0.857 \pm 0.022$\\ \hline
                f1-score\\ \hline
                $C_0: 1.000 \pm 0.000$\\
                $C_1: 0.913 \pm 0.013$\\
                $C_2: 0.902 \pm 0.011$\\
             \end{tabular}
             & - & - \\
    \end{tabular}
    \caption{ Experimental Results of the sector-related case achieved by the VAE and the Encoder (Enc). Columns report the 95\% confidence interval of precision, recall and f1-score w.r.t. the modes and the resolution actions of ATCO, for the non-weighted and weighted measures.}
    \label{tab:SR_results}
\end{table}

Table \ref{tab:SR_results} reports the 95\% confidence interval of the non-weighted and weighted versions of precision, recall and f1-score, achieved by the VAE and the Encoder (Enc) for the ATCO modes and the resolution actions, for the sector-related case. The structure of the table is similar to that of Table \ref{tab:SI_results}.

Regarding the modes of the ATCO reactions, the VAE network achieves f1-score of at least $0.835\pm0.025$ ($C_2$ mode) on all modes, for both the non-weighted and the weighted measures, while the Encoder achieves f1-score of at least $0.774\pm0.024$ ($C_2$ mode). VAE achieves the best results with a weighted f1-score of at least $0.945\pm0.015$ on all modes. The VAE model outperforms the Encoder model on all measures, weighted or not.

Similarly, for the sector-related case,  weighted measures are higher than the non-weighted. This shows that in many cases the model makes false predictions that are penalized lightly by the weighted measures, given that, as explained in the sector-ignorant case,  they are not critical.

As shown in Table \ref{tab:SR_results}, the non-weighted precision of mode $C_1$ is higher of that of mode $C_2$, while the non-weighted recall of mode $C_1$ is lower than that of mode $C_2$. On the other hand regarding the weighted measures the situation is quite the opposite, as the weighted precision of mode $C_1$ is lower of that of mode $C_2$, while the weighted recall of mode $C_1$ is higher than that of mode $C_2$. As explained in the sector-ignorant case, this implies that according to the non-weighted measures, there are cases where the model should assign a resolution action to a specific trajectory point by predicting mode $C_1$, but it does not, as in that particular point it instead predicts mode $C_2$. On the other hand, according to the weighted measures, the model predicts mode $C_1$ near the actual RATPs  even for points that are annotated as $C_2$.

Regarding the predictions of resolution actions, in the sector-related case results are not good: The f1-score of the $A_2$ resolution action is $0.384\pm0.075$ for the non-weighted and $0.419\pm0.076$ for the weighted measure. As already pointed out, this will be further explored in future work.

As CD\&R in the ATM domain is safety critical, we also report in Table \ref{tab:noRAprediction} the cases where the models fail to predict a resolution action and do not react at all in a critical situation. The ``setting" column reports the experimental setting, and columns ``VAE \#cases" and ``Encoder \#cases" report the the number of cases within the 95\% confidence interval for the VAE and the Encoder, respectively.

For the sector-ignorant setting we observe that the average number of cases for VAE is $6$, which is $3.79\%$ of the resolution actions in the test set. The Encoder on the other hand reports less cases where it did not react at all in a critical situation, with an average number of $3.2$ cases, which corresponds to $2.02\%$ of the resolution actions in the test set.
For the sector-related 
setting the number of such cases for VAE is $5.2$ corresponding to $7.6\%$ of the resolution actions in the test set. For the Encoder on the other hand the number of such cases is smaller, with an average value of $2.6$ cases, corresponding to $3.78\%$ of the resolution actions in the test set.

Considering the results reported in Tables \ref{tab:SI_results} and \ref{tab:SR_results}, in conjunction to the number of cases where the models did not react at all to critical situations reported in Table \ref{tab:noRAprediction}, we observe that the predictions of VAE, when compared with those of the Encoder, fit better the modes of the test set. Specifically, the window of points at which the VAE model predicts mode $C_1$ is more close to that of the expert, compared to the Encoder predictions. The Encoder on the other hand has less cases where it did not react at all to critical situations, compared to VAE. This, as we will discuss, could be explained by the probabilities that each model assigns to modes in each case.

Comparing the performance of the models between the different settings, we observe that models perform better for the sector-ignorant, rather than for the sector-related setting. This could be due to the difference of the size of the dataset between the two settings: The dataset for the sector-related case is approximately $1/3$ the size of the dataset for the sector-ignorant case. Another explanation for this difference is that flights from adjacent sectors may contribute to conflicts in the current sector (AoR) or in  the downstream sector of a flight. Such flights are not  considered in the sector-related case, therefore the corresponding conflicts are not detected.

\begin{table}[]
\centering
\small
    \begin{tabular}{c|c|c}
    Setting & VAE \#cases & Encoder \#cases \\ \hline
    sector-ignorant & $6.000\pm2.146$ & $3.200\pm1.746$ \\
    sector-related & $5.200 \pm 1.311$ & $2.600\pm1.022$ \\
    \end{tabular}
    \caption{ Number of cases within the 95\% confidence interval where the models do not predict a resolution action to any of the annotated or actual RATPS or any point in the time window of 70s near the actual or annotated RATPS.}
    \label{tab:noRAprediction}
\end{table}

\begin{table}[]
\tiny
    \begin{tabular}{c|c|c|c}
        Setting & Model & Expert & Predicted\\ \hline
        \rotatebox{90}{sector-ignorant} & \rotatebox{90}{VAE} & \includegraphics[width=0.38\linewidth]{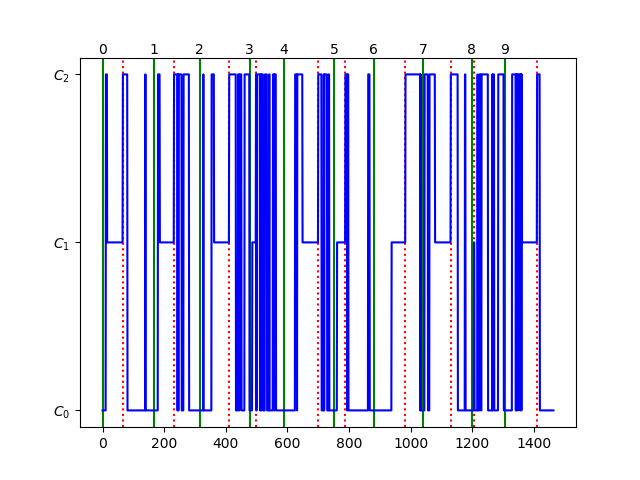} & \includegraphics[width=0.38\linewidth]{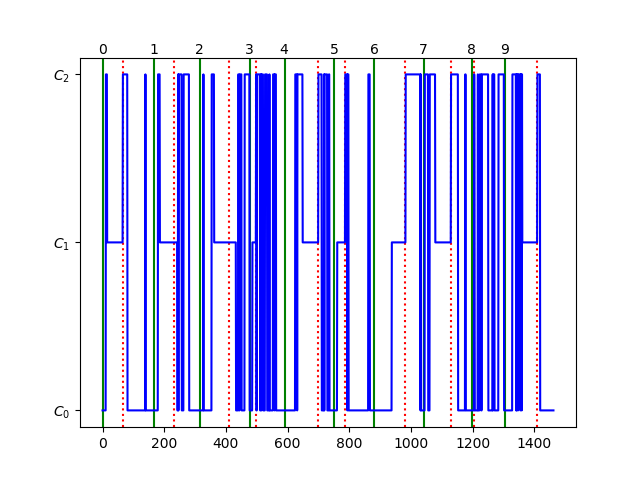} \\
        \hline
         \rotatebox{90}{sector-ignorant} & \rotatebox{90}{Enc} & \includegraphics[width=0.38\linewidth]{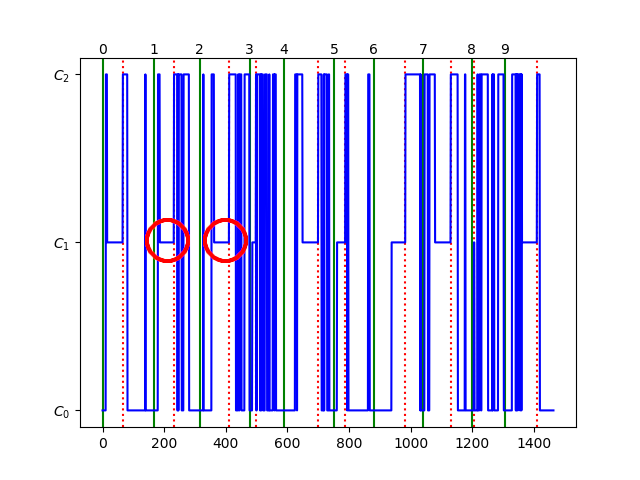} & \includegraphics[width=0.38\linewidth]{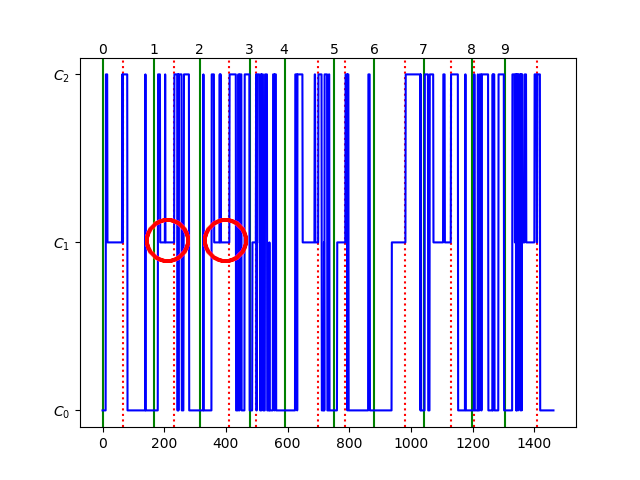} \\
        \hline
         \rotatebox{90}{sector-related} & \rotatebox{90}{VAE}& \includegraphics[width=0.38\linewidth]{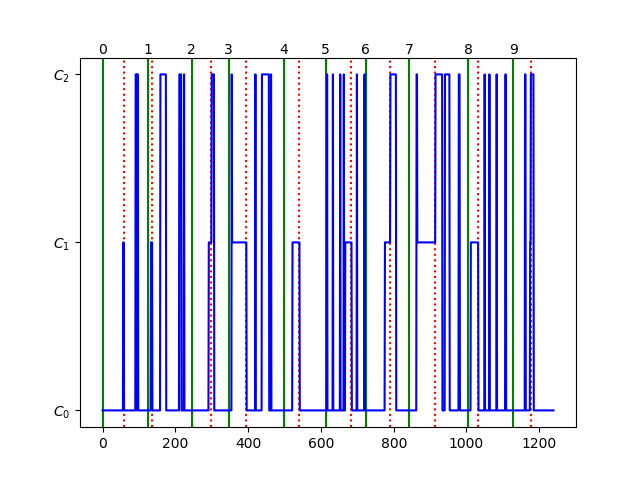} & \includegraphics[width=0.38\linewidth]{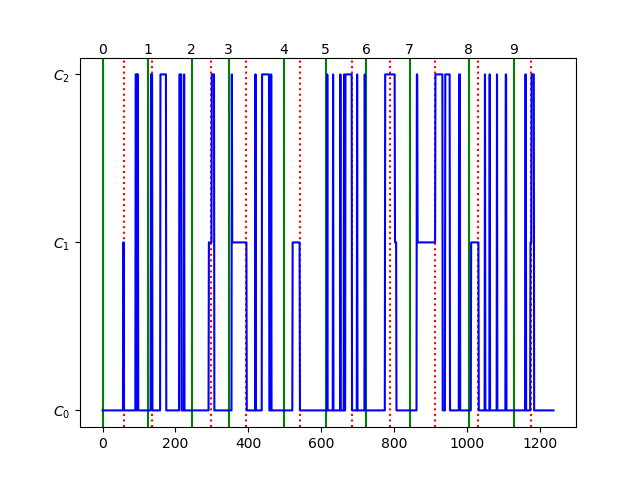} \\
        \hline
        \rotatebox{90}{sector-related} & \rotatebox{90}{Enc} & \includegraphics[width=0.38\linewidth]{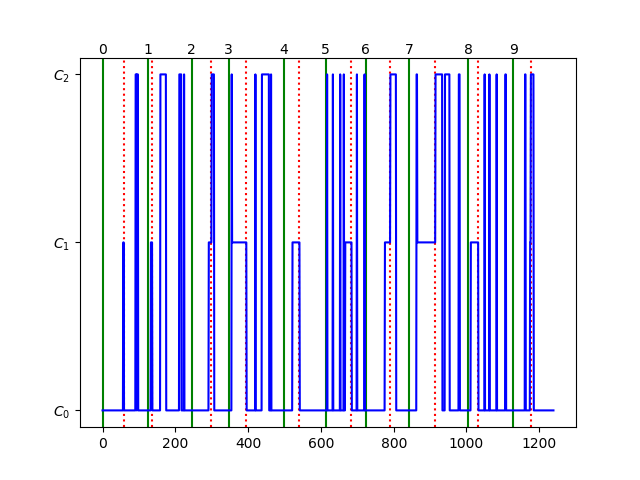} & \includegraphics[width=0.38\linewidth]{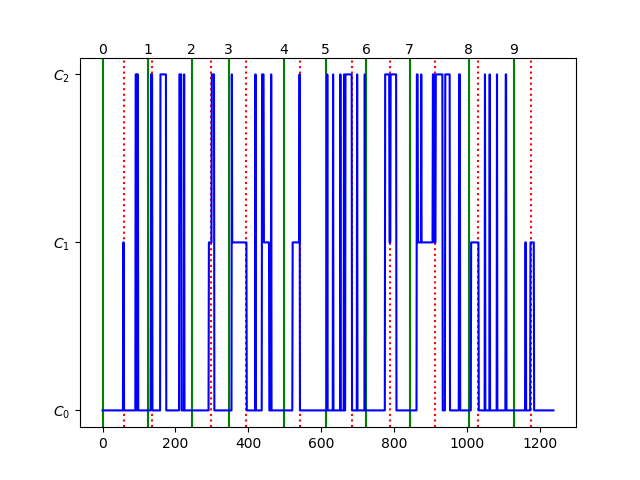} \\  
    \end{tabular}
    \caption{Modes according to the test set (column ``Expert") and the predictions of the models (column ``Predicted") for 10 trajectories.  X-axis: sequence number for all trajectories. Y-axis: modes. Green vertical lines show the start of each trajectory, while red dashed ones show the point of the actual RATP. Numbers over the green lines denote the sequence number of each trajectory. Red circles indicate examples of deviations between the predictions and the dataset.}
    \label{tab:modesFigs}
\end{table}

Delving deeper into the experimental results, Table \ref{tab:modesFigs} provides plots depicting the modes according to the test set (column ``Expert") and the predictions of the models (column ``Predicted") for  10 trajectories. Column ``Setting" denotes the experimental setting, and column ``Model" the model used for the predictions. In the x-axis we report the sequence number of the states of all trajectories and in the y-axis the modes. Green solid vertical lines denote the start of each trajectory, while red dashed ones denote the actual RATP. Numbers over the green lines denote the sequence number of each trajectory. We observe that for both experimental settings, both models accurately predict the modes with small deviations, as for instance predictions made by the Encoder on trajectories 1 and 2 for the sector-ignorant case (indicated by red circles).

Table \ref{tab:scatterplots} depicts the probability assigned in each mode by each model (VAE or Encoder (Enc)) at every point of the 10 trajectories. Column ``Setting" denotes the sector-ignorant or sector-related experimental setting. In the x-axis we report the sequence number of the states of all trajectories and in the y-axis the probability of each mode. Green solid vertical lines denote the start of each trajectory, while red dashed ones denote the actual RATP. Numbers over the green lines denote the sequence number of each trajectory. We observe that for the sector-ignorant case VAE provides more ``confident" predictions compared to the Encoder, assigning higher/lower probabilities to modes. For the sector-related case both models provide quite confident predictions and at most points the probabilities assigned to each mode are either high or low.

 The differences observed in the sector-ignorant case regarding the magnitude of the probabilities assigned to the different modes by each model, could explain why a) the predictions of the VAE are more consistent with the expert data compared to the predictions made by the Encoder, and b) the encoder has less cases where it did not react at all to critical situations, compared to VAE. VAE being more confident in its predictions, assigning higher/lower probabilities, will predict more consistently in a window of points one mode or the other, not changing easily between predicted modes from point to point. The Encoder on the other hand is assigning more mid-range probabilities to modes and could alternate more easily between modes predicted on consecutive points. Examples of such cases are indicated with the red circles in Table \ref{tab:modesFigs}. This implies that VAE when predicting a resolution action will be more committed to the prediction of the resolution action for a window of points, thus better fitting the dataset around the RATPs. On the other hand, the Encoder, being less "confident" and more "fluid" in its predictions, can predict at points mode $C_1$, even if these points are in windows of points where it mostly predicts mode $C_2$. This can result to less cases where it does not react at all to critical situations.
 
\begin{table}[]
\small
    \begin{tabular}{c|c|c}
        Setting & VAE & Enc\\ \hline
        \rotatebox{90}{sector-ignorant} & \includegraphics[width=0.4\linewidth]{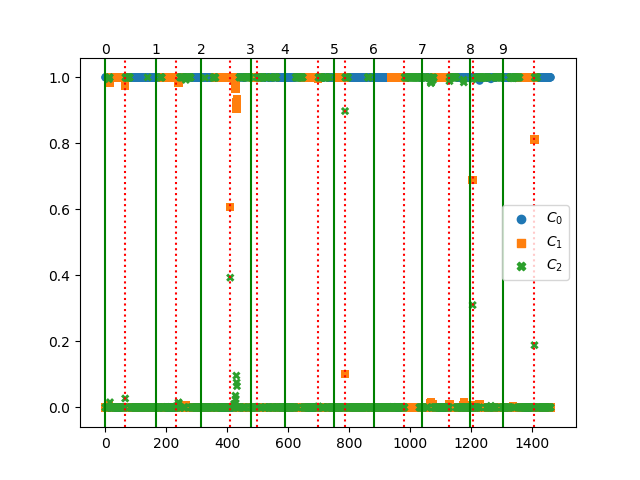} & \includegraphics[width=0.4\linewidth]{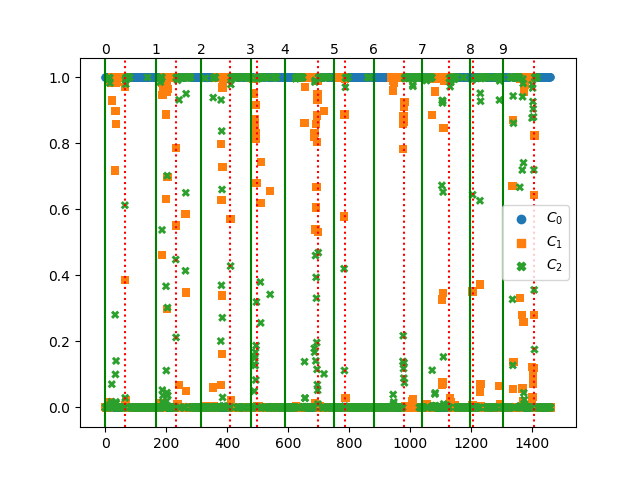}\\
        \hline
        \rotatebox{90}{sector-related} & \includegraphics[width=0.4\linewidth]{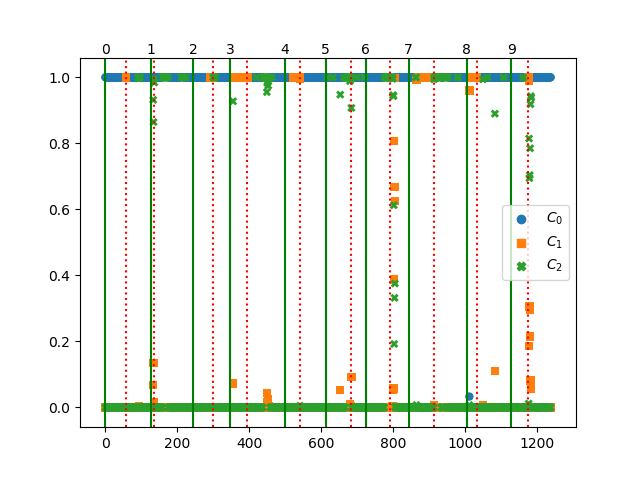} & \includegraphics[width=0.4\linewidth]{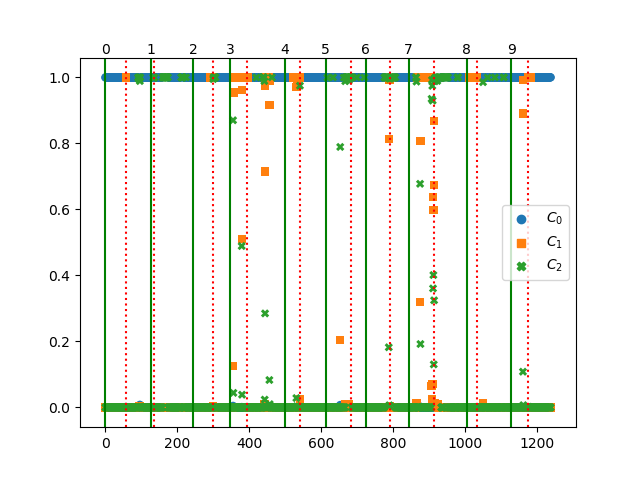} \\
        \hline
    \end{tabular}
    \caption{Scatterplots depicting the probability assigned at each mode by the VAE and the Encoder (Enc) at every point in 10 trajectories. X-axis: the sequence number of the states of all trajectories. Y-axis: the probability of each mode. Green vertical lines denote the start of each trajectory and red ones denote the actual RATP. Numbers over the green lines denote the state number of trajectories.}
    \label{tab:scatterplots}
\end{table}


\section{Related work}
\label{sec:related work}
In recent years multiple works consider the problem of assisting the ATCO with the CD\&R task. \cite{ribeiro2020review} provides a survey of CD\&R research both on manned and unmanned aviation. \cite{islami2017large} proposes a methodology to address strategic planning involving continent scale traffic. This method finds an optimal de-conflicted route and a departure time for each flight, relying on a hybrid-metaheuristic optimization algorithm that combines the advantages of simulated annealing and of hill-climbing local search methods. In \cite{dougui2013light} the authors propose a light propagation algorithm inspired by nature in order to generate conflict free 4-D trajectories, resolving conflicts at the tactical phase of operations. Also, in \cite{durand2006genetic} the authors propose a genetic algorithm based approach to the en-route conflict resolution problem at the tactical phase of operations. In \cite{srivatsatowards} authors use a lattice-based search space exploration AI planner to perform conflict resolution. In \cite{ayhan2018prescriptive} authors exploit an HMM to predict at the pre-tactical phase the evolution of trajectories based on historical trajectories and weather observations. The proposed method uses these predictions to detect conflicts and assign conflict-related probabilities to states. Resolution actions are decided by a variant of the Viterbi algorithm. 

The application of Reinforcement Learning (RL) methods on the CD\&R task has also received a lot of attention. Authors in \cite{pham2019reinforcement} and \cite{pham2019machine} explore a deep RL approach, based on Deep Deterministic Policy Gradient (DDPG) to resolve conflicts between two aircraft in the presence of uncertainty. While in \cite{dalmauair} authors formulate the problem as a multi-agent reinforcement learning problem and propose a Message Passing Actor Critic Model inspired by DGN \cite{jiang2018graph}, while also exploiting Message Passing Neural Networks \cite{gilmer2020message}. In \cite{ghosh2020deep} the authors combine Kernel Based RL with deep MARL to resolve conflicts by applying speed changes in real-time, also considering other factors such as fuel consumption and airspace congestion. Authors in \cite{isufajtowards} use Multi-Agent Deep Deterministic Policy Gradient (MADPG) to resolve conflicts, also considering time, fuel consumption and airspace complexity.

Closer to our approach are methods that somehow consider the ATCO preferences, either in a data-driven way as in \cite{calvo2017conflict}, \cite{tran2019intelligent} and \cite{van2020toward}, or by using rules and procedures derived from human experts as in \cite{erzberger2005automated}.

The work reported in \cite{calvo2017conflict} proposes a conflict resolution method operating at the strategic phase of operations. This method projects the aircraft's position into the future using the latest updated flight plans. 
The proposed methodology utilizes a data-driven model that a) classifies the conflict resolution maneuvers according to the relationship between the aircraft involved in the conflict and b) clusters the conflict resolution actions, considering the centroid of each cluster as a possible solution. Next, the method utilizes an $\epsilon$-constrained multi-objective optimization method to find the Pareto-optimal solutions w.r.t. the minimization of fuel consumption and the maximization of the likelihood of the resolution being implemented by an ATCO.


In \cite{tran2019intelligent} the authors propose a conflict resolution  advisory  system, able to  incorporate  human  preferences. The system uses an interactive  conflict  solver  (iCS)  for  acquiring  and 
characterizing human resolutions in conjunction to a RL  agent  that  learns  to  resolve  conflicts incorporating the characteristics of human resolution acquired by the iCS. This work  focuses on heading changes, deciding the trajectory change point (TCP). TCP is the point at which an aircraft after changing its heading to resolve a conflict will turn again towards its initial track. 

The method proposed in \cite{van2020toward} aims to provide personalized advisories to controllers. Authors train a convolutional NN on individual controller's data recorded from a human-in-the-loop simulation to predict  conflict resolution actions. The exploited dataset is in the form of Solution Space Diagrams (SSD), integrating various critical parameters of the CD\&R problem. 

Finally, \cite{erzberger2005automated}  describes an algorithm that provides 4D conflict resolution trajectories, based on a set of rules and procedures derived from human experts and from operational insights and analytical studies that reveal the characteristics of efficient conflict resolution techniques.

In this work we propose supervised deep learning techniques to learn models of ATCO reactions in resolving conflicts. This  implies  learning when the ATCO will react towards resolving a conflict, and which resolution action the ATCO will decide. In contrast to the other data-driven methods mentioned above, this work advances the state of the Art in CD\&R automation by exploiting recorded ATCO resolution actions on historical trajectories, formulating and  addressing the ATCO reaction problem, considering both abstract and low-level ATCO reactions, to imitate the ATCO. 

\section{Concluding remarks}
\label{sec:concluding remarks}
In this paper we have formulated the problem of conflict detection and resolution as a data-driven problem, aiming to learn ATCO reactions as a hierarchical task involving high-level reactions representing the mode of the ATCO behaviour, and low-level reactions representing ATCO resolution actions. We propose the use of a deep learning method employing a Variational Auto-Encoder (VAE) in the context of a deep learning methodology towards imitating ATCO, and have evaluated the proposed method using real world data in two different experimental settings: the sector-ignorant and the sector-related. To train the proposed model, we have developed a data-driven method for simulating the evolution of trajectories, incorporating uncertainty and revealing the conflicts that ATCO may have assessed before reacting. To evaluate the proposed method, as well as any other data-driven method that aims to solve the ATCO reaction prediction problem, we propose weighted precision, recall and F1 measures, and we use them to compare the VAE model against a basic model comprising  only an encoder. This comparison delves into the difference that the backwards propagation of the VAE decoder errors makes to the performance of VAE.

According to our experimental evaluation, both models (VAE and Encoder) accurately predict the mode of the ATCO behaviour in both experimental settings: the sector-ignorant and the sector-related. The VAE achieves consistently better results than the Encoder w.r.t. the weighted and non-weighted measures in both settings. The Encoder
on the other hand performs better w.r.t. the number of cases where the models do not predict a resolution action to any point near the actual RATP, or to any point in the time window of $70s$ near the annotated RATPS. These are cases where the model did not react at all to critical situations. This said, we must point out that the number of such cases for both models is very small. 

Regarding the predictions of resolution actions, results are not so impressive as those achieved on the prediction of modes, and this will be further explored in our future work,  refining the policy learned by the decoder.

Finally, regarding the two experimental settings, models perform better at the sector-ignorant case. This could be due to the difference of the size of the training dataset in the two settings, as the dataset for the sector-related case has approximately 1/3 of the size of the dataset for the sector-ignorant case.  Another explanation for this difference is that flights from adjacent sectors may contribute to conflicts in the current or downstream sector of a flight: These cases of conflict will not be detected in the sector related case.

Summarizing, the findings of this research are as follows:

-- Our methodology accurately predicts the modes of the ATCO behavior, predicting whether and when the ATCO reacts in the presence of conflicts.

-- According to the experimental evaluation the VAE model assigns high/low probabilities to modes, predicting one mode or the other without many fluctuations in predictions. The Encoder on the other hand assigns  more  mid-range  probabilities  to  modes,  and  fluctuates frequently between  modes  in  consecutive  trajectory points. As a result, the predictions of the VAE are more consistent with the expert reactions compared to the predictions made by the Encoder, although the Encoder has less cases where it did not react at all to critical situations, compared to VAE. The accuracy of predictions  implies that errors propagating backwards from the decoder to the Encoder play indeed an important role to the VAE model learned.

-- Predictions made on the low-level ATCO conflict resolution actions, were not as accurate as the predictions of the ATCO modes of reaction, and this will be further explored in future work. The decision regarding the resolution action depends on different factors, such as the evolution of the trajectories and evolution of conflicts during and after the performance of the resolution action, the preferences of ATCO, and also in some minor degree the preferences of the airspace users.

-- Regarding the different experimental settings explored, models perform better at the sector-ignorant case compared to the sector-related. This is something to be further explored in our future work. 

\section{Acknowledgements}
This research has received funding from the SESAR Joint Undertaking under the European Union’s Horizon 2020 Research and Innovation Programme under grant agreement No 783287. The opinions expressed herein reflect the authors’ view only. Under no circumstances shall the SESAR Joint Undertaking be responsible for any use that may be made of the information contained herein.





 \bibliographystyle{elsarticle-num} 
 \bibliography{cas-refs}





\end{document}